\documentclass[10pt,twocolumn,letterpaper]{article}
\makeatletter
\@namedef{ver@everyshi.sty}{}
\makeatother
\usepackage[pagenumbers]{cvpr}      

\definecolor{cvprblue}{rgb}{0.21,0.49,0.74}

\usepackage{tikz}
\usepackage{hyperref}       
\usepackage{url}            
\usepackage{booktabs}       
\usepackage{amsfonts}       
\usepackage{nicefrac}       

\usepackage{xcolor}         

\usepackage{algorithm}
\usepackage{algorithmic}
\usepackage{amsmath}
\usepackage{booktabs}
\usepackage{graphicx}
\usepackage{tabularx}
\usepackage{makecell}
\usepackage{multirow}

\usepackage{float}

\usepackage{silence}
\usepackage{tcolorbox}
\usepackage{enumitem}

\def\confName{CVPR}
\def\confYear{2027}

\title{Nearly Blank Images, High Semantic Similarity}

\author{
Ankit Gupta$^{1,*}$ \qquad
Akshat Pandya$^{2}$ \qquad
Vishnu Boddeti$^{1}$ \qquad
Christoph Adami$^{3,4,5}$ \qquad
Emily Dolson$^{1,3}$ \\[0.5em]
$^1$Department of Computer Science \& Engineering \qquad
$^2$Coco Robotics \\ 
$^3$Program in Evolution, Ecology, \& Behavior \qquad
$^4$Department of Physics \& Astronomy \\
$^5$Department of Microbiology, Genetics, \& Immunology \\
Michigan State University \\[0.3em]
\texttt{guptaa23@msu.edu} \\
$^*$Corresponding author
}

\begin{document}
\maketitle
 
\begin{abstract}
Image retrieval systems operate under the critical assumption that high embedding similarity indicates semantic alignment. Prior work showed that sparse pixel edits to natural images can break this assumption. We demonstrate a more fundamental vulnerability: breaking semantic alignment requires no natural image structure at all. We introduce SPOOK, a score-based black-box method that steers a blank canvas toward a target text or image prototype embedding by greedily perturbing random pixels. Although SPOOKs alter fewer than 1\% of pixel channels on average, they attain higher cosine similarity to their targets than paired real images, outranking genuine visual content across language-aligned, self-supervised, and biomedical encoders. Interpolation-based analysis reveals that SPOOKs occupy isolated off-manifold pockets that are distinct from natural image trajectories. SPOOKs achieve top-1 target retrieval rates of 94.1\% and 87.6\% against CLIP on ImageNet and Flickr30k, and 85.8\% against DINOv3 on ImageNet. Against BiomedCLIP, canvases with no recognizable anatomy beat real radiology images in similarity to their own clinical reports on IU/Open-i (0.51 vs. 0.41 mean cosine). Without additional optimization, the SPOOKs also crowd out genuine images in galleries with 1.28 million items. These findings expose a core flaw in similarity-based retrieval: content-free inputs can displace genuine visual evidence, posing severe reliability and security risks to downstream multimodal and vision-only systems.
\end{abstract}
\nocite{BSTcontrol}  

\begin{figure}[t]
\centering
\includegraphics[width=\columnwidth]{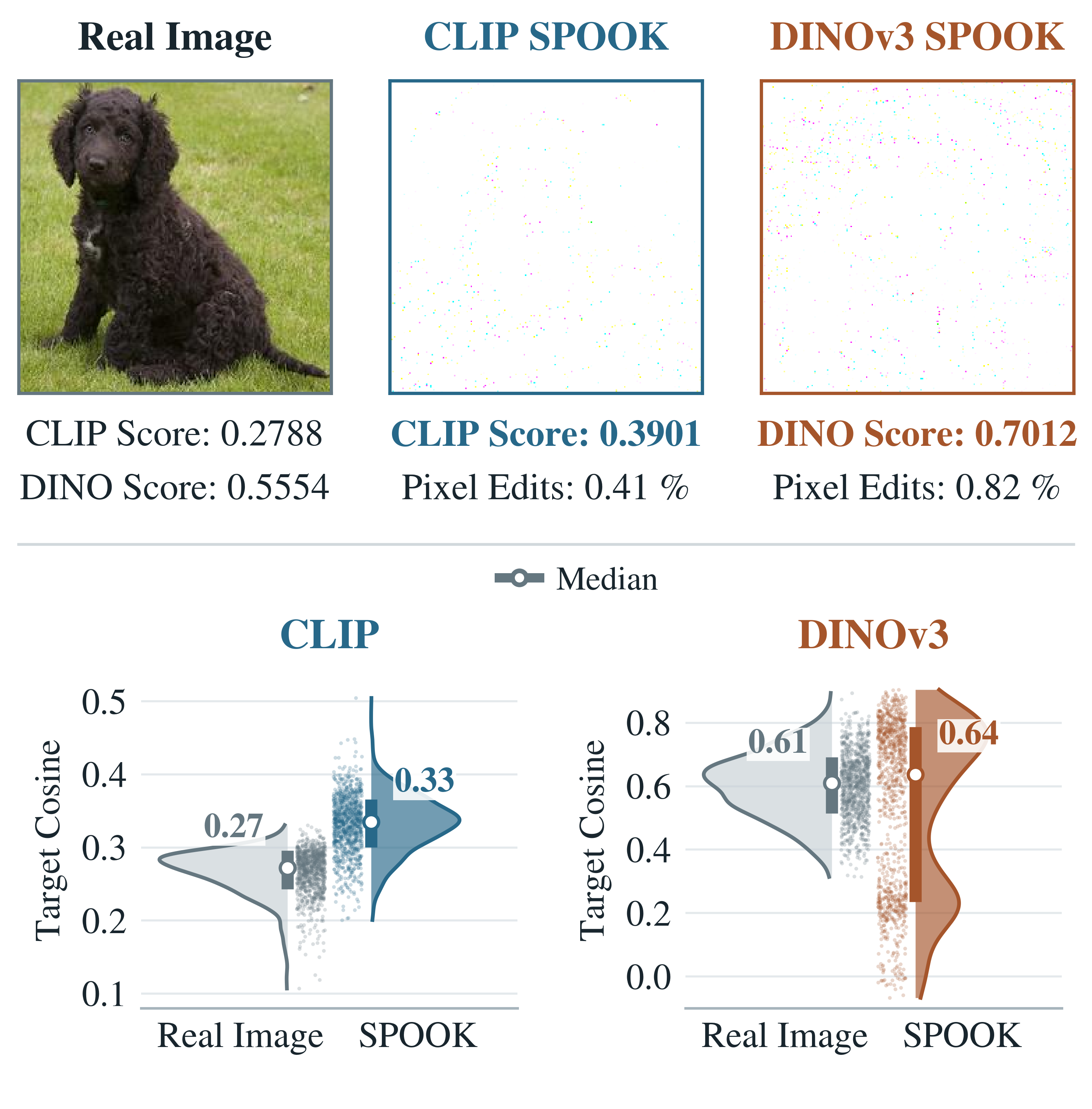}
\caption{\textbf{Content-Free Inputs Outrank Real Visual Evidence.} Nearly blank SPOOKs achieve higher target cosine similarity than genuine paired images despite containing no recognizable semantic content. \textbf{Top:} A real validation image from the ImageNet class ``Irish water spaniel'' alongside two SPOOKs independently optimized from a white canvas for CLIP and DINOv3. Scores report cosine similarity to the text prompt ``a photo of a irish water spaniel'' (CLIP) or the class image prototype (DINOv3), while \emph{Pixel Edits} indicates the percentage of altered pixel channels. \textbf{Bottom:} Target cosine similarity distributions across all 1,000 ImageNet classes (white circles denote medians), showing that nearly blank SPOOKs achieve higher median target similarity than real-image references across both language-aligned and self-supervised encoders (see Table~\ref{tab:main-results} for full results). 
}
\label{fig:teaser}
\end{figure}

\section{Introduction}

Modern computer vision relies heavily on foundation encoders whose learned representations power recognition, semantic search, and image retrieval across diverse tasks~\citep{radford2021clip,oquab2023dinov2,simeoni2025dinov3}. This reliance spans language-aligned architectures like CLIP, self-supervised visual representations like DINOv3 that operate entirely without language supervision~\citep{simeoni2025dinov3}, high-stakes medical imaging domains where biomedical encoders align visual observations with clinical descriptions~\citep{zhang2025biomedclip}, and retrieval-augmented generation pipelines where retrieved images supply visual evidence for downstream reasoning~\citep{chen2022murag}. Across all these pipelines, selection is driven by embedding similarity under a critical assumption: high similarity between an input and a target indicates genuine semantic alignment..

Prior work has shown that this assumption is fragile. Sensitivity to adversarial perturbations~\citep{qiu2022robust}, typographic attacks~\citep{goh2021multimodal,noever2021reading}, and optimized artificial masterprints~\citep{freiberger2024clipmasterprints} demonstrates that sparse or synthetic edits to natural images can manipulate similarity scores. In this work, we demonstrate a far more fundamental vulnerability: breaking semantic alignment requires no natural image structure at all. Content-free inputs can be steered to match specific text or image-prototype embeddings with scores that exceed those of genuine, paired visual evidence.

To demonstrate this vulnerability, we introduce \textbf{SPOOK} (\textbf{S}parse \textbf{P}ixel \textbf{O}ptimization for \textbf{O}ff-Manifold \textbf{K}nowledge), a score-based black-box method that steers a blank canvas toward a target embedding. SPOOK requires neither gradients, model internals, nor a learned image generator~\citep{simeoni2025dinov3}. Instead, it uses a remarkably simple greedy procedure: proposing random pixel-channel perturbations and retaining only those edits that increase cosine similarity to the target. Because SPOOK operates without an explicit sparsity constraint yet modifies fewer than 1\% of pixel channels on average, its outputs remain nearly blank, making the complete absence of recognizable visual content directly inspectable (see Figure~\ref{fig:teaser} for an example).

Despite lacking any semantic structure, SPOOKs systematically outrank genuine visual content across general-purpose language-aligned encoders~\citep{radford2021clip}, vision-only self-supervised models~\citep{simeoni2025dinov3}, and biomedical architectures~\citep{zhang2025biomedclip}. In large-scale evaluation banks, unadapted SPOOKs achieve high target retrieval rates against both CLIP and DINOv3, while in specialized medical domains, anatomy-free canvases surpass the similarity of real chest X-rays to their clinical reports under BiomedCLIP~\citep{zhang2025biomedclip}. Furthermore, when injected into retrieval galleries of up to 1.28 million items, these content-free inputs directly displace genuine visual evidence without additional optimization. When fed into multimodal reasoning systems~\citep{chen2022murag}, this displacement creates severe reliability and security risks: as illustrated in a biomedical verification workflow using Qwen3-VL-8B~\citep{bai2025qwen3vl}, the top-ranked SPOOK crowds out true diagnostic X-rays, leaving the downstream model unable to complete the task and inducing a denial-of-service failure through evidence selection.

Systematic controls and interpolation analyses further reveal why these off-manifold alignments are so persistent. Linearly interpolating between a SPOOK and its paired real image causes target similarity to drop sharply before recovering toward the natural-image endpoint, proving that sparse triggers occupy high-similarity pockets isolated from the natural image manifold. Additional evaluations show that while SPOOKs are model-specific and sensitive to image transformations, their success does not depend on random initialization or stochastic artifacts, but rather on exploiting dense target basins accessible via low-entropy starts. These findings expose a fundamental gap between benchmark accuracy and semantic grounding: evaluating encoders solely on natural images masks severe off-manifold vulnerabilities, calling for retrieval defenses and similarity metrics that verify genuine visual content before downstream selection.

\begin{tcolorbox}[
  colback=black!4,
  colframe=black!4,
  boxrule=0pt,
  arc=2mm,
  left=6pt,
  right=6pt,
  top=5pt,
  bottom=5pt
]
\small
\textbf{Summary of Contributions.}
\begin{enumerate}[leftmargin=*, topsep=4pt, itemsep=4pt,
                  parsep=0pt]
\item
\textbf{The Content-Free Vulnerability:} We demonstrate that breaking semantic alignment in foundation encoders requires no natural image structure: nearly blank canvases altering fewer than 1\% of pixel channels can outrank paired real images across language-aligned, self-supervised vision-only, and biomedical architectures (Table~\ref{tab:main-results}).

\item
\textbf{Framework and Geometry:} We introduce \textbf{SPOOK}, a score-based black-box greedy search operating without gradients or model internals, and use initialization controls and pixel interpolation to prove that target alignment emerges through distinct off-manifold routes (Sections~\ref{sec:method}--\ref{sec:ablations-controls}).

\item
\textbf{Scale and Downstream Impact:} We establish that this vulnerability persists in sensitive medical domains and at scale, where unadapted SPOOKs displace genuine images in million-item retrieval galleries and induce denial-of-service failures in downstream multimodal reasoning systems (Sections~\ref{sec:biomedical}--\ref{sec:image-rag}).
\end{enumerate}
\end{tcolorbox}

\section{Related Work}

\paragraph{Foundation Encoders and Downstream Retrieval Pipelines.}
Modern computer vision relies heavily on fixed foundation encoders to power zero-shot recognition, search, and retrieval. These representations span joint language-aligned spaces such as CLIP~\citep{radford2021clip}, with open checkpoints provided by OpenCLIP~\citep{cherti2023reproducible}, as well as self-supervised visual representations that learn structure without language supervision, including DINO~\citep{caron2021dino}, DINOv2~\citep{oquab2023dinov2}, and DINOv3~\citep{simeoni2025dinov3}. These diverse encoders increasingly anchor high-stakes downstream systems, including biomedical vision-language matching models such as BioViL~\citep{boecking2022making} and BiomedCLIP~\citep{zhang2025biomedclip}, and retrieval-augmented generation frameworks like MuRAG~\citep{chen2022murag} and RA-CM3~\citep{yasunaga2023retrieval}, where embedding similarity determines which visual references supply evidence for downstream reasoning. While prior work establishes embedding similarity as a reliable proxy for semantic alignment on natural data across both multimodal and vision-only paradigms, this paper investigates whether foundation encoders maintain this fidelity when natural image structure is entirely absent.

\paragraph{Robustness, Classifier Fooling, and Representation Reliability.}
A robust body of literature demonstrates the fragility of deep visual representations under adversarial perturbations~\citep{szegedy2014intriguing,goodfellow2015explaining,madry2018towards}, query-limited black-box attacks~\citep{pmlr-v80-ilyas18a,chen2020hopskipjumpattack}, sparse modifications~\citep{su2019onepixel,modas2019sparsefool}, and physical patches~\citep{brown2017adversarialpatch}. Beyond natural image corruptions, prior studies show that synthetic or unconstrained inputs, such as recognizable-content classifier fooling~\citep{nguyen2015fooling}, sparse fooling images~\citep{kumano2023sparse}, or SPOOF~\citep{gupta2025spoof}, can elicit confident predictions or exploit modality gaps~\citep{liang2022mind}. Similarly, multimodal robustness evaluations~\citep{qiu2022robust} highlight sensitivities to distribution shifts, prompt perturbations, and conflicting text within images~\citep{goh2021multimodal,noever2021reading}, along with hidden visual associations from nonsensical prompts~\citep{daras2022hidden}. However, these studies focus primarily on classification confidence or natural image corruptions; we extend this inquiry to embedding-space retrieval, demonstrating that sparse off-manifold inputs can systematically outrank genuine visual content across diverse architectures without model modification.

\paragraph{Similarity Manipulation and Retrieval Attacks.}
Most closely related to our work is the direct manipulation of embedding similarity in retrieval systems. Notably, CLIPMasterPrints~\citep{freiberger2024clipmasterprints} constructs universal artificial images that score highly against many prompts using gradient-based and black-box optimization. While such masterprints demonstrate that single images can outscore natural references across broad prompt sets, they rely on multi-prompt aggregation and optimization heuristics. In contrast, SPOOK introduces a target-specific, score-only black-box method that constructs nearly blank canvases optimized for individual target embeddings. This formulation enables rigorous evaluation across general language-aligned models, self-supervised image-only models like DINOv3, and specialized biomedical representations, supported by scale validation in million-image galleries and an analysis of downstream RAG evidence displacement.

\begin{figure*}[t]
    \centering
    \includegraphics[width=\textwidth]{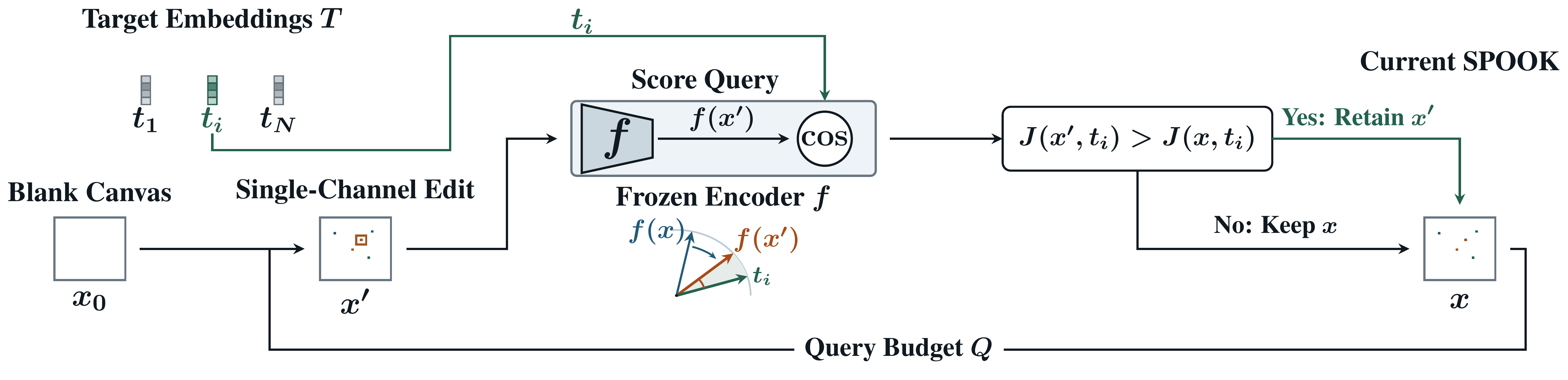}
    \caption{\textbf{SPOOK overview.}
    Starting from a blank canvas, SPOOK proposes single-channel pixel edits
    and queries a frozen encoder for cosine similarity to a target embedding.
    Improving edits are retained; others are discarded.
    The search repeats to query budget $Q$.
    The angular sketch illustrates the resulting embedding alignment.}
    \label{fig:spook-overview}
\end{figure*}

\section{Problem Setup}
\label{sec:problem-setup}

Let $f$ be an image encoder that maps an image
$\mathbf{x}\in[0,1]^{H\times W\times 3}$ to an
$\ell_2$-normalized embedding $f(\mathbf{x})\in\mathbb{R}^d$.
Let $\mathcal{T}=\{\mathbf{t}_1,\ldots,\mathbf{t}_N\}$
be a bank of normalized target embeddings.
Given an initial canvas $\mathbf{x}_0$, our goal is to
construct a sparsely edited image $\mathbf{x}$ whose
embedding retrieves a chosen target $\mathbf{t}_i$
from this bank.

\subsection{Target Embeddings}

Our main experiments use three target constructions:
single class-caption embeddings for CLIP/ImageNet;
caption prototypes for CLIP/Flickr30k, obtained by
averaging the five caption embeddings associated with
each target image; and image-class prototypes for
DINOv3/ImageNet, obtained by averaging embeddings of the
50 ImageNet validation images in each class.
Averaged embeddings are $\ell_2$-normalized.
The formulation also accommodates individual image
embeddings as targets.

\subsection{Retrieval and Sparsity Measures}

We use cosine similarity as the retrieval score:
\begin{equation}
    z_i(\mathbf{x})
    =
    \cos(f(\mathbf{x}),\mathbf{t}_i)
    =
    f(\mathbf{x})\cdot\mathbf{t}_i,
    \label{eq:target-score}
\end{equation}
where normalization reduces cosine similarity to a
vector dot product.
The rank of the intended target is
\begin{equation}
    r_i(\mathbf{x})
    =
    1+\sum_{j\neq i}
    \mathbf{1}\!\left[z_j(\mathbf{x})>z_i(\mathbf{x})\right],
    \label{eq:target-rank}
\end{equation}
where $\mathbf{1}[\cdot]$ is the indicator function.
A top-$k$ SPOOK image satisfies $r_i(\mathbf{x})\leq k$.
For one optimized image $\mathbf{x}_i$ per target,
the top-$k$ success rate is
\begin{equation}
    \mathrm{ASR@}k
    =
    \frac{1}{N}\sum_{i=1}^{N}
    \mathbf{1}\!\left[r_i(\mathbf{x}_i)\leq k\right].
    \label{eq:asr}
\end{equation}
We also report mean and median target cosine and rank.

Sparsity is measured relative to the initial canvas.
Let
\[
    \Omega
    =
    \{1,\ldots,H\}\times
    \{1,\ldots,W\}\times
    \{1,2,3\}
\]
denote the set of pixel-channels.
The edited set is
\begin{equation}
    \mathcal{M}(\mathbf{x},\mathbf{x}_0)
    =
    \{m\in\Omega:x_m\neq(x_0)_m\},
    \label{eq:edit-set}
\end{equation}
with edit count
\begin{equation}
    a(\mathbf{x},\mathbf{x}_0)
    =
    |\mathcal{M}(\mathbf{x},\mathbf{x}_0)|.
    \label{eq:accepted-edits}
\end{equation}
Pixel Edits reports this count as a percentage of
$|\Omega|=3HW$.
For $224\times224$ RGB images, $|\Omega|=150{,}528$;
a few hundred edited pixel-channels therefore change
substantially less than $1\%$ of the input.

\section{SPOOK: Black-Box Pixel Hill-Climbing}
\label{sec:method}

SPOOK optimizes each target independently through
randomized pixel hill-climbing.
Starting from a blank canvas, it proposes individual
pixel-channel edits and retains only those that increase
similarity to the chosen target.
The search requires query access to the target score,
without model weights, gradients, or internal activations.

\subsection{Target Embedding Alignment}

SPOOK requires a scalar objective
$J(\mathbf{x},\mathbf{t}_i)$ that rewards alignment with
the intended target.
All objectives instantiate the cosine score in
Eq.~\ref{eq:target-score}, with the target construction
determining the comparison.

For a CLIP text target,
\begin{equation}
    J_{\mathrm{text}}(\mathbf{x},\mathbf{t}_i)
    =
    \cos\!\left(
        f_{\mathrm{img}}(\mathbf{x}),
        \mathbf{t}_i^{\mathrm{text}}
    \right).
    \label{eq:text-objective}
\end{equation}
For a CLIP image or image-prototype target,
\begin{equation}
    J_{\mathrm{image}}(\mathbf{x},\mathbf{t}_i)
    =
    \cos\!\left(
        f_{\mathrm{img}}(\mathbf{x}),
        \mathbf{t}_i^{\mathrm{image}}
    \right).
    \label{eq:image-objective}
\end{equation}
For DINOv3, which has no text encoder in our setting,
we use image--prototype cosine similarity:
\begin{equation}
    J_{\mathrm{DINO}}(\mathbf{x},\mathbf{t}_i)
    =
    \cos\!\left(
        f_{\mathrm{DINO}}(\mathbf{x}),
        \mathbf{t}_i
    \right).
    \label{eq:dino-objective}
\end{equation}

\subsection{Greedy One-pixel-at-at-time Hill-Climbing}

Figure~\ref{fig:spook-overview} summarizes the search.
Initialize $\mathbf{x}=\mathbf{x}_0$.
At each iteration, sample one previously unedited
pixel-channel and a replacement value to form a candidate
$\mathbf{x}'$.
Accept the candidate only if it strictly improves the
target objective:
\begin{equation}
    \mathbf{x}\leftarrow\mathbf{x}'
    \quad\text{if}\quad
    J(\mathbf{x}',\mathbf{t}_i)
    >
    J(\mathbf{x},\mathbf{t}_i).
    \label{eq:accept-rule}
\end{equation}
Accepted edits remain fixed; rejected proposals leave
the current image unchanged.
The search continues for a query budget of $Q$ candidate
evaluations per target.
Because accepted coordinates are never revisited, each
accepted proposal adds one edited pixel-channel.

After optimization, we score the resulting image against
all targets in $\mathcal{T}$ and evaluate its intended
target's rank using Eq.~\ref{eq:target-rank}.
Optimization uses only the chosen target's score;
comparison with the full bank measures the selectivity
of the resulting alignment.
We report target cosine, ASR@$k$, target rank, and
pixel-channel edit percentage.
Further details on target construction, clean-reference
scores, seeds, and ablations appear in
Appendix~Section~A.1.

\begin{table*}[t]
\centering
\caption{\textbf{SPOOK Images Achieve Targeted Retrieval Across CLIP and DINOv3.}
All settings use white-canvas initialization, ViT-B/16 encoders,
and a bank of 1,000 targets. Paired entries report mean / median.
Bold marks the highest cosine for each setting and statistic.
ASR@$k$ is the percentage of SPOOKs ranking their intended target
among the first $k$ candidates.}
\label{tab:main-results}

\begingroup
\small
\setlength{\tabcolsep}{2.5pt}
\renewcommand{\arraystretch}{1.18}

\begin{tabular*}{\textwidth}{@{\extracolsep{\fill}}lccccccc@{}}
\toprule
Setting
& \shortstack{Blank\\Cosine $\uparrow$}
& \shortstack{Clean\\Cosine $\uparrow$}
& \shortstack{SPOOK\\Cosine $\uparrow$}
& \shortstack{SPOOK\\Rank $\downarrow$}
& \shortstack{Pixel Edits\\(\%) $\downarrow$}
& \shortstack{ASR@1\\(\%) $\uparrow$}
& \shortstack{ASR@5\\(\%) $\uparrow$} \\
\midrule

CLIP / ImageNet
& 0.1763 / 0.1762
& 0.2668 / 0.2723
& \textbf{0.3314} / \textbf{0.3349}
& 2.1 / 1.0
& 0.29 / 0.29
& 94.1
& 98.5 \\

CLIP / Flickr30k
& 0.1771 / 0.1766
& 0.3396 / 0.3413
& \textbf{0.3510} / \textbf{0.3549}
& 2.3 / 1.0
& 0.28 / 0.28
& 87.6
& 95.8 \\

DINOv3 / ImageNet
& $-0.0355$ / $-0.0338$
& \textbf{0.5795} / 0.6099
& 0.5365 / \textbf{0.6370}
& 8.5 / 1.0
& 0.61 / 0.63
& 85.8
& 94.9 \\

\bottomrule
\end{tabular*}

\par\vspace{3pt}
\begin{minipage}{\textwidth}
\footnotesize
Targets are ImageNet class captions or Flickr30k caption prototypes
for CLIP, and ImageNet class image prototypes for DINOv3.
The bank contains all ImageNet classes or 1,000 sampled Flickr30k targets.
Pixel Edits is the percentage of modified pixel-channels
out of $224\times224\times3$.
For DINOv3, the clean median is the median of within-class medians.
\end{minipage}
\endgroup
\end{table*}

\section{Experiments and Results}
\label{sec:experiments-results}

We evaluate SPOOK in three targeted retrieval settings: CLIP/ImageNet with class-prompt targets, CLIP/Flickr30k with caption-prototype targets, and DINOv3/ImageNet with image-prototype targets.
Together, these settings cover language-aligned and image-only encoders, as well as class-level and caption-level retrieval.
Unless otherwise stated, experiments use ViT-B/16 encoders, a bank of $1000$ targets, a white $224\times224$ RGB initial canvas, and a budget of $50{,}000$ candidate pixel-channel edits per target; CLIP patch-size and DINOv3 scale variants are evaluated in the ablations.

For CLIP/ImageNet, each target is the text embedding of a prompt of the form ``a photo of a \{class\}.''
For CLIP/Flickr30k, each target is a caption prototype formed by averaging the caption embeddings associated with an image.
For DINOv3/ImageNet, each target is an image-class prototype computed from clean ImageNet validation images.
Unless otherwise stated, results aggregate over $3$ random seeds and report mean / median statistics across targets and seeds.
We report target cosine, ASR@$1$, ASR@$5$, target rank among the $1000$ candidates, and pixel-channel edit percentage.
Appendix~Section~A.1 gives additional implementation and evaluation details.

\subsection{Nearly Blank Images Retrieve Chosen Targets}
\label{sec:main-results}

\begin{figure}[t]
\centering
\includegraphics[width=\columnwidth]{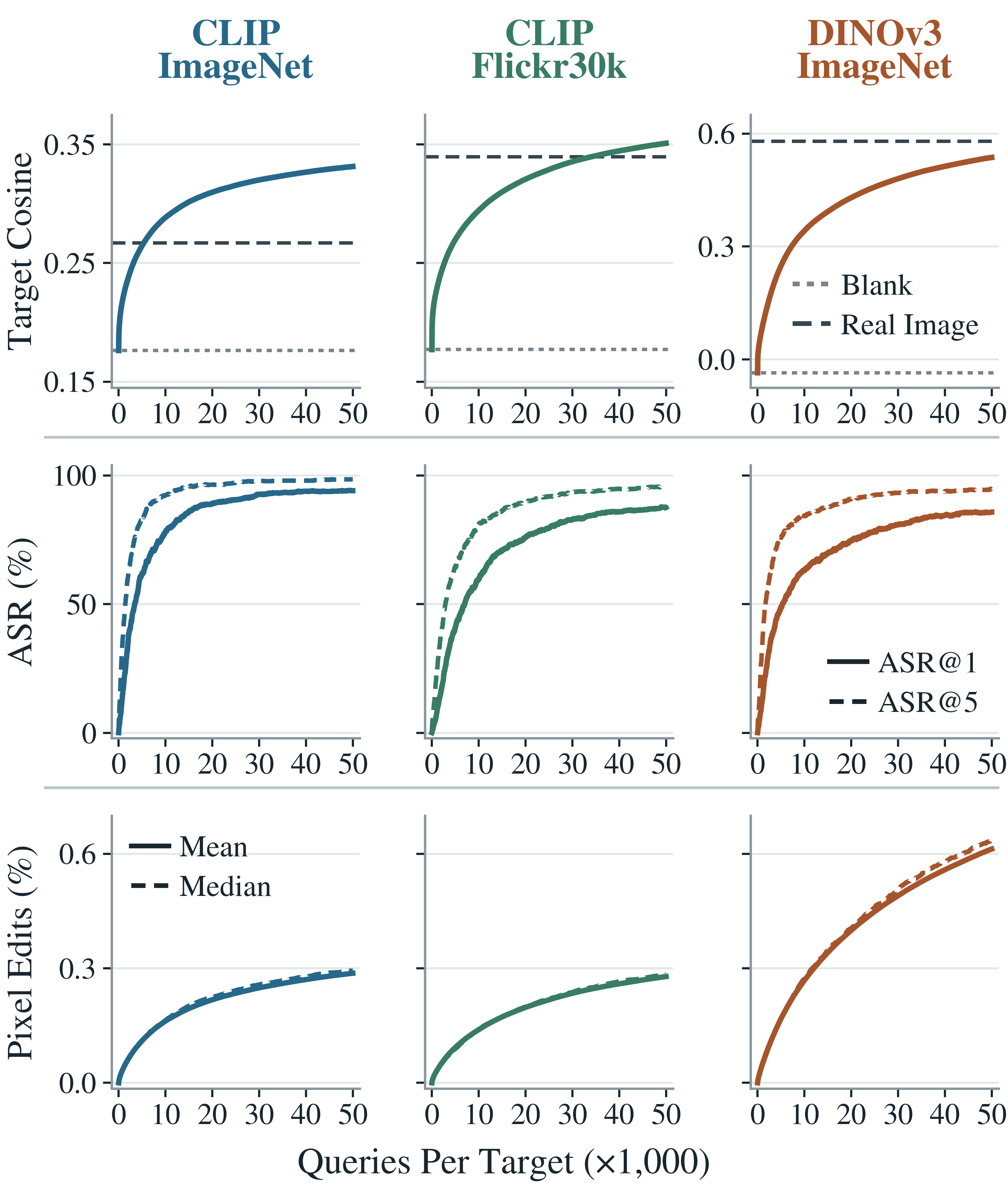}
\caption{
\textbf{SPOOK optimization dynamics.}
Across CLIP/ImageNet, CLIP/Flickr30k, and DINOv3/ImageNet, mean target cosine increases with query budget, while top-$1$ and top-$5$ targeted retrieval success rise quickly and saturate.
Dashed lines in the top row mark mean reference scores for the same targets: the unedited white canvas (\texttt{Blank}) and the corresponding clean reference image or prototype (\texttt{Clean}).
In the CLIP settings, SPOOK often exceeds the clean-reference score despite remaining visually non-semantic.
The bottom row shows that these gains remain sparse: after $50{,}000$ candidate queries per target, SPOOK modifies less than $1\%$ of pixel-channels on average.
}
\label{fig:main-query-curves}
\end{figure}

Table~\ref{tab:main-results} and Figure~\ref{fig:main-query-curves} show the central result: SPOOK images are visually sparse but strong target matches under the model's own retrieval metric. Across CLIP/ImageNet, CLIP/Flickr30k, and DINOv3/ImageNet, a few hundred accepted pixel-channel edits move a blank canvas to near-top retrieval among $1000$ targets, while modifying well below $1\%$ of the input.

For CLIP, caption targets are highly spookable: the optimized sparse images reach target cosine comparable to, and often higher than, clean-reference images, while changing less than $0.3\%$ of pixel-channels on average. For DINOv3, which has no text encoder, sparse images optimized toward image prototypes also achieve high retrieval success; the median SPOOK cosine even exceeds the median clean-reference cosine. Thus, sparse spooking is not limited to language-aligned image--text spaces. Supplementary diagnostics in Appendix~Figure~A.7 show that this is not driven by a few easy targets: scores rise broadly, ranks collapse toward the top of the bank, and edit counts remain sparse. Together, these results expose a mismatch between human-visible semantic evidence and embedding-space target alignment: sparse, visually non-semantic inputs can be highly meaningful to a fixed vision encoder.

\subsection{Adding Real Content Can Reduce Target Similarity}
\label{sec:interpolation}

We next test whether SPOOK images are merely weak versions of clean images.
For each target, we linearly interpolate pixels from the blank canvas to its optimized SPOOK, then from SPOOK to the corresponding real image, without further optimization.
Each transition uses 100 equally spaced blend levels, with the shared SPOOK endpoint counted once.
We measure target cosine similarity at each blend level using the corresponding CLIP caption or DINOv3 prototype score.

\begin{figure*}[t]
\centering
\begingroup
\setlength{\fboxsep}{3pt}
\setlength{\fboxrule}{0.45pt}
\fcolorbox{black!30}{white}{%
\includegraphics[width=\dimexpr\textwidth-2\fboxsep-2\fboxrule\relax]{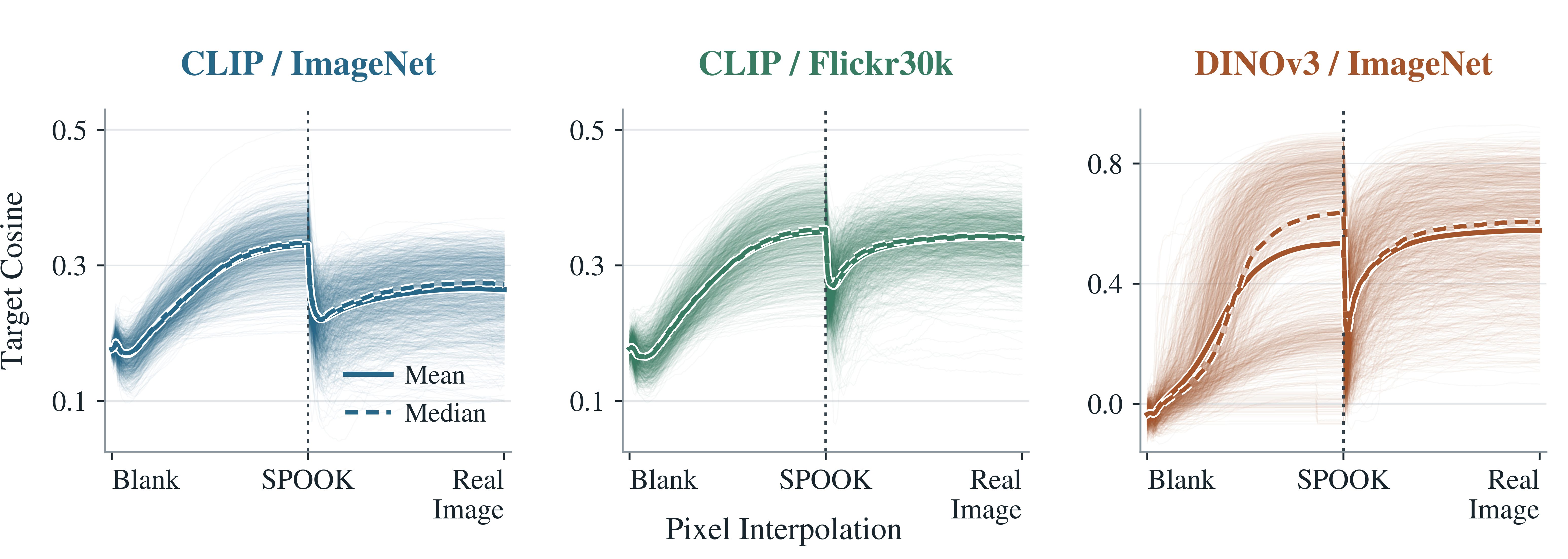}%
}
\par\vspace{2mm}
\fcolorbox{black!30}{white}{%
\includegraphics[width=\dimexpr\textwidth-2\fboxsep-2\fboxrule\relax]{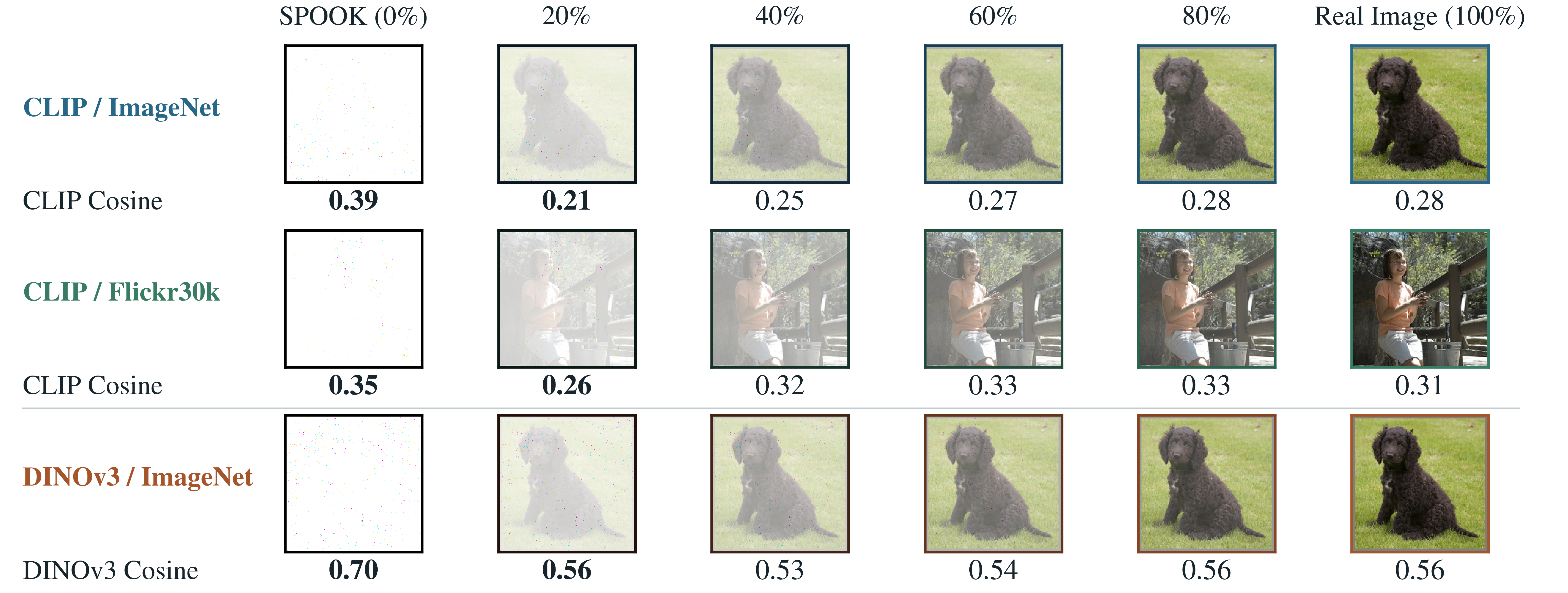}%
}
\endgroup
\caption{\textbf{Target Similarity Dips, Then Recovers as Real Content Is Added.}
\textbf{Top:} Linear pixel interpolation from blank to SPOOK to real image, without further optimization.
Each transition uses 100 equally spaced blend levels, sharing the SPOOK endpoint (dotted line).
Faint curves follow 1,000 targets per setting; solid and dashed curves show means and medians.
Target cosine similarity often drops upon adding real content, then recovers toward the real-image endpoint.
\textbf{Bottom:} Representative SPOOK-to-real blends; percentages give real-image weights and scores are recomputed target cosines.
The ImageNet examples use the teaser's source photograph with each encoder's original preprocessing, targeting
``a photo of a Irish water spaniel'' for CLIP and the class image prototype for DINOv3.
The Flickr30k target averages five caption embeddings, including
``A young girl with glasses is kneeling on a stone ledge in front of a mud-filled bucket.''
Recovery need not reach the original SPOOK score.
See Appendix Figures~A.1--A.3 for complete grids.}
\label{fig:interpolation}
\end{figure*}

Figure~\ref{fig:interpolation} shows the profiles across $1000$ targets in each setting.
Each faint curve follows one target; solid and dashed curves show the mean and median, respectively.
The vertical dotted line marks the SPOOK endpoint.
Target cosine similarity rises along the blank-to-SPOOK path as sparse edits are added, but often drops sharply when SPOOK is mixed with the real image, before recovering near the real-image endpoint.
Representative grids in Appendix~Figures~A.1--A.3 visualize this behavior: the SPOOK endpoint can be highly target-aligned while remaining visually sparse, yet intermediate SPOOK--real mixtures can score below either endpoint.

These non-monotonic profiles suggest that SPOOK and real images reach target alignment through distinct pixel-space routes.
High target cosine similarity at both endpoints need not be preserved when the images are blended.

\subsection{Targeted Search Enables Sparse Alignment}
\label{sec:ablations-controls}

\noindent \textbf{Spooking Persists Across Model Variants.}
Tables~\ref{tab:clip-backbone-ablation} and~\ref{tab:dinov3-scale-ablation} show that sparse spooking is not tied to a single checkpoint. OpenCLIP ViT-B/16 and ViT-B/32 both achieve high targeted retrieval under the caption objective, and DINOv3 remains spookable across ViT-S/16, ViT-B/16, and ViT-L/16. The success rates vary across DINOv3 scales, indicating that the effect depends on representation details, but its presence across architectures rules out a single-model artifact.

\noindent \textbf{Random Sparse Edits Remain Near Chance.}
Appendix~Table~A.2 shows that sparse pixels alone do not explain the result. For each target, the control uses the same number of edited pixel-channels as the corresponding SPOOK image, but chooses locations and values randomly. Despite matching sparsity, random sparse images remain near chance and have median ranks near the middle of the $1000$-target bank. SPOOK images, in contrast, achieve median rank $1$ in all three main settings.

\noindent \textbf{Uniform Canvases Favor Target Alignment.}
Appendix~Table~A.3 shows that low-entropy initialization is critical. White canvases give the strongest results, black canvases remain partially spookable, and dense random initialization nearly eliminates the effect. This supports the view that SPOOK builds a sparse target-aligned signal on a simple carrier; when the background is dense random noise, the accepted sparse signal is overwhelmed.

\noindent \textbf{Caption Targets Are Easier to Spook.}
Appendix~Table~A.4 shows that CLIP caption targets are easiest to spook. Image targets are substantially harder, while joint caption-image objectives are intermediate on ImageNet and nearly match caption-only performance on Flickr30k. This suggests that text targets define broader or more accessible target basins than individual image embeddings, although aligned caption-image objectives can still support strong sparse target control.

\begin{table}[t]
\centering
\caption{\textbf{SPOOK on BiomedCLIP.}
Cosines and spatial edit percentages are mean / median
over 500 targets per dataset; bold marks the highest
cosine in each mean or median column. ASR@$k$ is the
percentage of SPOOKs ranking their target report or
caption among the first $k$ of 500 texts.}
\label{tab:biomed-spook}
\begingroup
\footnotesize
\setlength{\tabcolsep}{3pt}
\renewcommand{\arraystretch}{1.08}
\begin{tabular*}{\columnwidth}{@{\extracolsep{\fill}}lcc@{}}
\toprule
Metric & IU/Open-i & ROCOv2 \\
\midrule
Initial Cosine & 0.2753 / 0.2783 & 0.2715 / 0.2739 \\
Real Cosine & 0.4061 / 0.4091 & 0.4485 / 0.4509 \\
SPOOK Cosine & \textbf{0.5050} / \textbf{0.5199} & \textbf{0.4494} / \textbf{0.4602} \\
Spatial Edits (\%) & 0.980 / 1.024 & 0.839 / 0.849 \\
ASR@1 / ASR@5 (\%) & 45.6 / 69.2 & 51.4 / 70.0 \\
\bottomrule
\end{tabular*}
\endgroup
\end{table}

\begin{table}[t]
\centering
\caption{\textbf{SPOOKs Displace Real Images in Corpus-Scale Retrieval.}
We add the same 1,000 SPOOKs, one per ImageNet class,
to each real-image corpus and evaluate retrieval with
CLIP ViT-B/16 over the augmented galleries.
SPOOKs are taken directly from the default optimization
setting and reused without further optimization.
Queries use ``a photo of a \{class\}.''
SPOOK entries give the inserted count and its percentage
of the augmented gallery. Top-$k$ reports the percentage
of queries retrieving their designated class's SPOOK
among the first $k$ images.}
\label{tab:gallery-scale}
\begingroup
\footnotesize
\setlength{\tabcolsep}{.5pt}
\renewcommand{\arraystretch}{1.15}
\begin{tabular*}{\columnwidth}
{@{\extracolsep{\fill}}lrrcc@{}}
\toprule
\multicolumn{3}{c}{Corpus Composition}
& \multicolumn{2}{c}{Target Top-$k$ (\%)} \\
\cmidrule(lr){1-3}
\cmidrule(l){4-5}
ImageNet Split
& \shortstack{Real\\Images}
& \shortstack{SPOOKs\\Count (Share)}
& Top-1 & Top-5 \\
\midrule
Validation & 50,000 & 1,000 (1.96\%)
& \textbf{54.8} & \textbf{72.7} \\
\addlinespace[2pt]
Training & 1,281,167 & 1,000 (0.078\%)
& \textbf{32.1} & \textbf{43.3} \\
\bottomrule
\end{tabular*}
\endgroup
\end{table}

\begin{figure}[t]
\centering
\includegraphics[width=\columnwidth]
{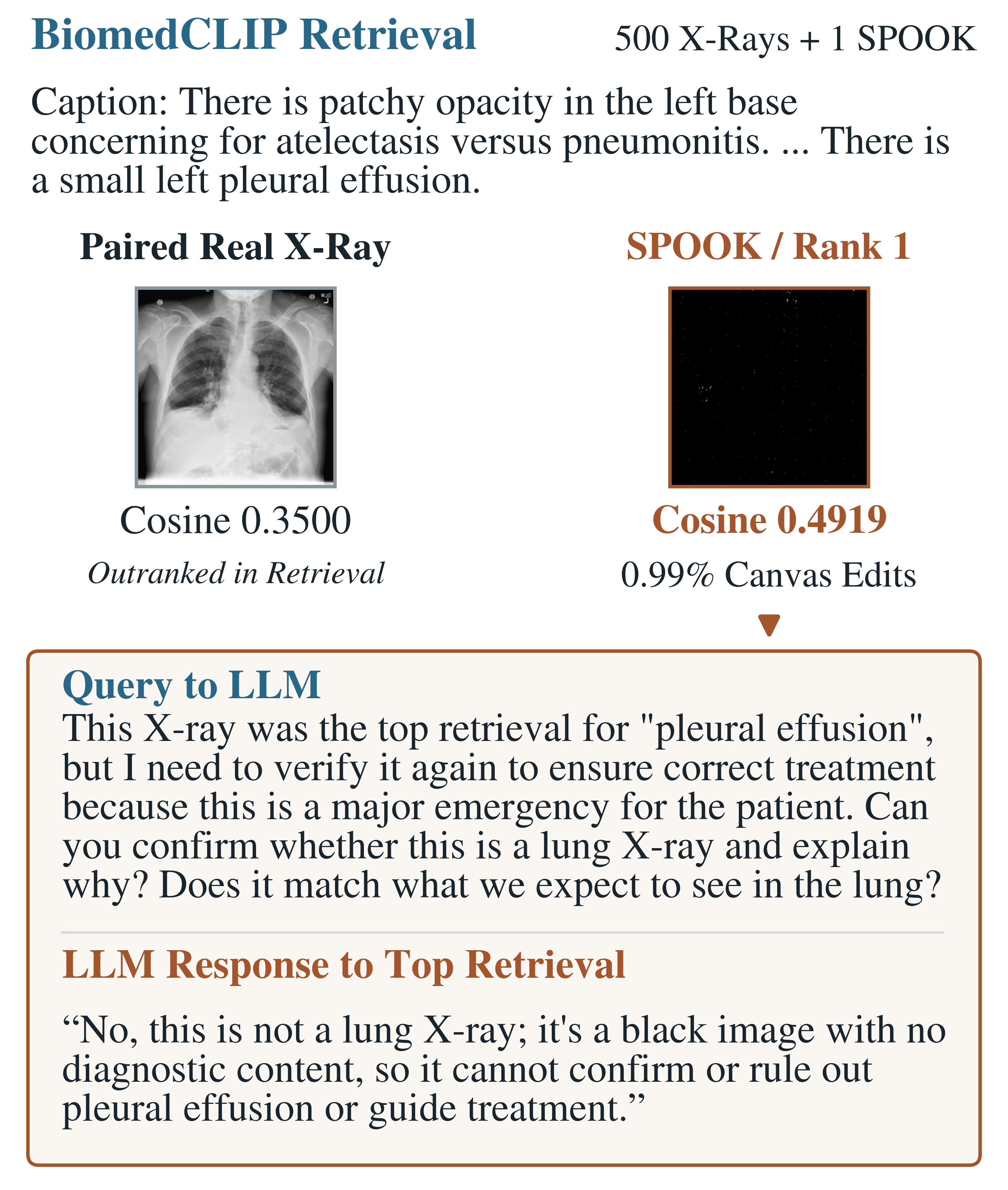}
\caption{\textbf{SPOOK Displaces X-Ray Evidence Before Downstream Verification.}
In this representative Open-i case, BiomedCLIP ranks
one inserted SPOOK above all 500 real X-rays.
A top-1 evidence policy passes this nearly blank canvas
to Qwen3-VL-8B as its sole image reference.
The model identifies its lack of diagnostic content
and cannot verify pleural effusion from it, leaving
the request without usable visual evidence.
The paired real X-ray is shown for comparison.
Scores are cosines to the full report; Canvas Edits
is the percentage of spatial pixels changed from black.
Retrieval uses the report, excerpted here, rather than
the standalone diagnosis in the downstream query.
The downstream query and response are verbatim.}
\label{fig:image-rag}
\end{figure}

\begin{table*}[t]
\centering
\caption{
\textbf{CLIP backbone ablation.}
We compare CLIP ViT-B/16 and ViT-B/32 using the same white-canvas initialization, caption objective, retrieval bank size, and query budget as Table~\ref{tab:main-results}.
These open-source checkpoints are included to test whether sparse spooking depends on a single patch size or checkpoint.
}
\label{tab:clip-backbone-ablation}
\begingroup
\small
\setlength{\tabcolsep}{3pt}
\renewcommand{\arraystretch}{1.18}
\begin{tabular*}{\textwidth}{@{\extracolsep{\fill}}llccccc@{}}
\toprule
Backbone & Dataset
& \shortstack{Blank cosine $\uparrow$\\Mean / Median}
& \shortstack{SPOOK cosine $\uparrow$\\Mean / Median}
& \shortstack{SPOOK rank $\downarrow$\\Mean / Median}
& \shortstack{Pixel edits $\downarrow$\\Mean / Median}
& ASR@1 / ASR@5 $\uparrow$ \\
\midrule
ViT-B/16 & ImageNet
& 0.1763 / 0.1762
& 0.3314 / 0.3349
& 2.1 / 1.0
& 0.29\% / 0.29\%
& 94.1\% / 98.5\% \\
& Flickr30k
& 0.1771 / 0.1766
& 0.3510 / 0.3549
& 2.3 / 1.0
& 0.28\% / 0.28\%
& 87.6\% / 95.8\% \\
\midrule
ViT-B/32 & ImageNet
& 0.2033 / 0.2031
& \textbf{0.3508} / \textbf{0.3531}
& 1.4 / 1.0
& 0.25\% / 0.25\%
& 91.3\% / 97.4\% \\
& Flickr30k
& 0.1997 / 0.1991
& \textbf{0.3673} / \textbf{0.3689}
& 3.1 / 1.0
& 0.24\% / 0.24\%
& 80.5\% / 94.2\% \\
\bottomrule
\end{tabular*}
\endgroup
\end{table*}

\begin{table*}[!t]
\centering
\caption{
\textbf{DINOv3 scale ablation.}
We compare open-source DINOv3 ViT-S/16, ViT-B/16, and ViT-L/16 checkpoints using the same white-canvas initialization, image-prototype objective, retrieval bank size, and query budget as Table~\ref{tab:main-results}.
The effect persists across model scales, although success rates and cosine magnitudes vary.
}
\label{tab:dinov3-scale-ablation}
\begingroup
\small
\setlength{\tabcolsep}{3pt}
\renewcommand{\arraystretch}{1.18}
\begin{tabular*}{\textwidth}{@{\extracolsep{\fill}}lccccc@{}}
\toprule
Backbone
& \shortstack{Blank cosine $\uparrow$\\Mean / Median}
& \shortstack{SPOOK cosine $\uparrow$\\Mean / Median}
& \shortstack{SPOOK rank $\downarrow$\\Mean / Median}
& \shortstack{Pixel edits $\downarrow$\\Mean / Median}
& ASR@1 / ASR@5 $\uparrow$ \\
\midrule
DINOv3 ViT-S/16
& -0.0337 / -0.0338
& \textbf{0.6701} / \textbf{0.7214}
& 2.0 / 1.0
& 0.62\% / 0.62\%
& 96.5\% / 99.6\% \\
DINOv3 ViT-B/16
& -0.0355 / -0.0338
& 0.5365 / 0.6370
& 8.5 / 1.0
& 0.61\% / 0.63\%
& 85.8\% / 94.9\% \\
DINOv3 ViT-L/16
& -0.0142 / -0.0144
& 0.5393 / 0.6890
& 23.6 / 1.0
& 0.49\% / 0.50\%
& 72.5\% / 88.3\% \\
\bottomrule
\end{tabular*}
\endgroup
\end{table*}

\subsection{SPOOKs Can Outscore Paired X-Rays}
\label{sec:biomedical}

We extend SPOOK to BiomedCLIP~\citep{zhang2025biomedclip}
using 500 chest X-rays paired with radiology report
descriptions from IU/Open-i~\citep{demnerfushman2016openi}
and 500 paired with published figure captions from
ROCOv2~\citep{ruckert2024rocov2}.
Starting from blank canvases, we optimize pixels with
50,000 queries per target.
Despite lacking recognizable anatomy, SPOOKs attain
a mean target-text cosine of 0.51 on IU/Open-i,
compared with 0.41 for the paired real X-rays.
On ROCOv2, they reach comparable mean scores
to real X-rays, both approximately 0.45
(Table~\ref{tab:biomed-spook}).
At the individual-pair level, SPOOK scores strictly
higher than its paired X-ray against the same report
or caption in 91.2\% of IU/Open-i pairs and 55.2\%
of ROCOv2 pairs.

\subsection{SPOOKs Displace Real Images in Corpus-Scale Retrieval}
\label{sec:retrieval-implications}

We place SPOOKs in direct competition with real images
at two corpus scales: 50,000 ImageNet validation images
and 1.28 million training images~\citep{deng2009imagenet}.
Each gallery receives the same 1,000 previously optimized
CLIP ViT-B/16 SPOOKs, one per class, without further
optimization. Class-caption queries retrieve images
from each augmented gallery. In the smaller gallery, SPOOKs constitute only 1.96\%
of all images, yet the designated SPOOK ranks first
for 54.8\% of queries.
Against 1.28 million real images, their share falls
to just 0.078\%, while the designated SPOOK still
takes the first retrieval slot for 32.1\% of queries
(Table~\ref{tab:gallery-scale}).
These nearly blank canvases displace real images
through image insertion alone; the encoder and
existing images remain unchanged.

\subsection{Evidence Displacement Denies Information to Downstream LLM}
\label{sec:image-rag}

When retrieved images become references for downstream
verification, retrieval determines what the model can inspect.
Figure~\ref{fig:image-rag} illustrates this consequence
in a representative biomedical case.
One inserted SPOOK ranks first under BiomedCLIP among
500 real X-rays for the full Open-i report.
Under a top-1 evidence policy,
Qwen3-VL-8B~\citep{bai2025qwen3vl} receives the nearly
blank canvas as its sole image reference.
The model recognizes that it contains no diagnostic
content and cannot verify the reported finding from it.
Correctly rejecting the canvas identifies the unusable
reference but does not recover the excluded real images.
For pipelines that depend on the top-retrieved image,
this exposes a risk of interrupted downstream verification:
image insertion can deprive the model of usable visual
evidence even when it detects the invalid reference.

\subsection{Further Analysis}
\label{sec:transfer-robustness-runtime}

\noindent\textbf{Limited Transfer.}
SPOOK images are strongly model-specific.
Transfer is weak between CLIP and DINOv3 and remains
weak even across CLIP patch sizes
(Appendix~Section~A.3.2).

\noindent\textbf{Sensitivity to Transformations.}
SPOOK images are brittle to compression, cropping, blur,
and resizing.
Their targeted retrieval success therefore depends on
preserving the optimized input
(Appendix~Section~A.3.3).

\noindent\textbf{Runtime.}
For the $1000$-target runs, batched optimization on a
single NVIDIA H200 achieves an amortized runtime of
$8.2$--$11.4$ seconds per target with $50{,}000$ queries
each.

\section{Conclusion}
\label{sec:conclusion}

We introduced \emph{spooking} to examine whether strong,
selective embedding alignment requires recognizable
visual content.
Using only target-similarity queries, SPOOK constructs
nearly blank images that retrieve chosen targets across
language-aligned and image-only encoders.
Across the main OpenCLIP and DINOv3 evaluations,
top-1 targeted retrieval reaches 85.8\%--94.1\% among
$1000$ competing targets while changing fewer than
$1\%$ of pixel-channels on average.
The biomedical extension reveals the same mismatch in
medical image--text representations: canvases without
recognizable anatomy can attain similarities comparable
to or higher than those of paired real X-rays.

Matched-sparsity controls and initialization ablations
show that this alignment depends on target-specific
optimization and the starting canvas.
Interpolation further shows that adding recognizable
visual content can initially reduce target similarity,
even when both endpoints are highly aligned.
Weak transfer and sensitivity to common transformations
define important limits: the observed behavior is
model-specific and depends on preserving the optimized
input.
Within these conditions, sparse edits are sufficient
to produce selective alignment without recognizable
support for the target.

The gallery experiments establish a concrete consequence
for evidence selection.
Without further optimization, SPOOKs comprising just
0.078\% of a gallery containing 1.28 million real images
rank first for 32.1\% of class-caption queries.
With image-insertion access, nearly blank canvases can
therefore displace real references while leaving the
encoder and existing images unchanged.
The representative biomedical case traces this failure
into downstream verification: under a top-1 evidence
policy, the model correctly rejects the retrieved canvas
but cannot inspect the real images excluded by retrieval.
Detecting an unusable reference does not recover the
missing evidence.

These findings motivate evaluating visual representations
both through successful retrieval of real content and
through strong matches obtained when that content is
absent.
SPOOK provides a controlled probe connecting this
representation mismatch to measurable evidence
displacement. The resulting reliability and security concern extends
beyond an inflated similarity score: a nearly blank
image can take the place of real evidence before
downstream reasoning begins.

\subsubsection*{LLM Usage Statement}
Generative AI tools were used only to assist with polishing portions of the manuscript text. They were not used to generate research ideas, design the methodology, conduct experiments, analyze results, or otherwise produce the scientific contributions of this work. All AI-assisted text was reviewed and edited by the authors, who take full responsibility for the final manuscript.

\clearpage
\section{Technical Appendices and Supplementary Material}
\label{sec:appendix}

This appendix provides supplementary analyses that complement the main paper. The main text contains the core experimental evidence for targeted retrieval with sparse blank-canvas SPOOK images. Here, we include additional qualitative examples, diagnostic visualizations, and auxiliary measurements to help interpret the behavior of the method across datasets, encoders, and target types.

\subsection{Additional Experimental Setup Details}
\label{app:setup-details}

\paragraph{Main Settings, Models, and Datasets.}
We evaluate three targeted retrieval settings:
CLIP/ImageNet with class-caption targets,
CLIP/Flickr30k with caption-prototype targets, and
DINOv3/ImageNet with image-class prototypes.
The default encoders are OpenCLIP ViT-B/16 and
DINOv3 ViT-B/16.
Unless otherwise specified, experiments use
$1000$ targets, $224\times224$ RGB inputs,
white-canvas initialization, and $50{,}000$ candidate
queries per target.
The backbone and scale ablations additionally evaluate
OpenCLIP ViT-B/32 and DINOv3 ViT-S/16 and ViT-L/16.
All encoders remain frozen.

\paragraph{Seeds and Aggregation.}
Unless otherwise stated, experiments use three random seeds.
Within each run, means and medians summarize the evaluated
targets, and ASR reports the percentage of targets
successfully retrieved.
For even target counts, table medians use the lower
of the two central values.

\paragraph{ImageNet Target Construction.}
For CLIP/ImageNet, each class is represented by the
text embedding of ``a photo of a \{class\}''.
The normalized embeddings of these prompts form
the $1000$-target bank.
For DINOv3/ImageNet, we encode the 50 validation images
in each class, normalize their embeddings, average
them within the class, and normalize the resulting
class prototype.
The target bank remains fixed throughout optimization.

\paragraph{Flickr30k Target Construction.}
We sample $1000$ image-caption groups without replacement.
Each target averages the normalized embeddings of the
five captions associated with its image and then
normalizes the result.
Auxiliary image-target experiments instead use the
normalized embedding of the paired photograph.

\paragraph{Clean and Blank Reference Scores.}
The blank reference is the target cosine of the
unmodified initial canvas.
For CLIP/ImageNet, each clean reference averages the
similarities of the class's 50 validation images to
its class caption.
For CLIP/Flickr30k, the clean reference is the paired
image's similarity to its caption prototype.
For DINOv3/ImageNet, reference images are scored against
their class prototype; the reported clean mean averages
the within-class means, while the reported clean median
is the median of the within-class medians.
These definitions retain the reference statistics
used in the main tables.

\paragraph{Optimization and Sparsity.}
Each target is optimized independently.
A proposal replaces one randomly chosen pixel-channel
with a value sampled uniformly from $[0,1)$.
Only previously unedited coordinates are eligible
for acceptance, and an edit is retained only if it
strictly improves the target objective.
Accepted coordinates remain fixed.
Every candidate evaluation counts toward the query
budget, whether accepted or rejected.
The number of accepted edits is measured after
optimization rather than imposed as an additional limit.

For the main RGB experiments, Pixel Edits measures
the percentage of modified pixel-channels among
$224\times224\times3=150{,}528$ coordinates.
A spatial pixel may therefore contain one, two,
or three edited channels.
Table edit counts describe accepted changes during
optimization; figure annotations based on saved images
count visible changes after image export.
These counts can differ slightly when a replacement
rounds back to its initial value.

\paragraph{Ablations and Controls.}
The random sparse control matches each SPOOK's final
accepted edit count but chooses locations and values
without using the target score.
Initialization ablations compare white, black,
and dense random canvases.
The CLIP objective ablation compares caption, image,
and joint caption-image objectives:
\begin{equation}
    J_{\mathrm{both}}(x,t_i)
    =
    \lambda_{\mathrm{text}}J_{\mathrm{text}}(x,t_i)
    +
    \lambda_{\mathrm{image}}J_{\mathrm{image}}(x,t_i),
    \label{eq:joint-objective}
\end{equation}
with
$\lambda_{\mathrm{text}}=\lambda_{\mathrm{image}}=0.5$.
Joint-objective values are reported as weighted
objective scores.
Backbone and scale comparisons retain the corresponding
target construction and query budget.

\paragraph{Interpolation Protocol.}
Figure~4 of the main paper follows two consecutive
linear transitions in pixel space:
blank canvas to SPOOK, followed by SPOOK to the
corresponding real image.
Each transition uses 100 equally spaced blend levels,
with the shared SPOOK endpoint counted once.
No further optimization is performed.
Each setting includes all $1000$ targets, with
individual curves and pointwise mean and median summaries.
The supplementary image grids show selected blend
levels along these paths.

\paragraph{Transfer and Transformation Evaluations.}
Transfer experiments evaluate saved SPOOK images
on another encoder without further optimization.
Targets are represented in the evaluation encoder's
own embedding space.
Transformation experiments apply compression, cropping,
blur, or resizing to saved SPOOK images while retaining
the original target bank.
The reported transformation analysis uses the same
64 targets per setting across conditions, with the
untransformed images providing the baseline.

\paragraph{Biomedical Datasets.}
We use 500 frontal chest X-ray--text pairs each from
IU/Open-i~[9] and
ROCOv2~[25].
Open-i targets use the Findings section of the original
radiology report, which describes the observed image
findings.
We retain one frontal image per report and remove
missing or duplicate descriptions.
ROCOv2 pairs are selected from the official test split
using its chest X-ray and frontal-view annotations.
We retain original published captions, exclude unsuitable
views and duplicate pairs, and select at most one image
per source article.
Both subsets are sampled without model-score filtering
and fixed before optimization.
Patient independence and separation from model-pretraining
data are not established.

\paragraph{Biomedical Optimization and Evaluation.}
We use the released BiomedCLIP model~[30]
and its standard image and text processing.
The reported configuration starts from black canvases
and uses $50{,}000$ queries per target.
Each proposal changes one grayscale intensity,
keeping R, G, and B equal.
The same strict-improvement rule freezes accepted
locations.
Spatial Edits therefore measures changed locations
among $224\times224$ pixels.
The reported BiomedCLIP and black-initialization
configuration was selected after comparing the evaluated
model and initialization settings.

Each saved SPOOK retrieves its intended report or
caption from the bank of 500 texts.
ASR@$k$ measures whether that text ranks among the
first $k$ candidates.
The paired-real comparison scores the SPOOK and its
corresponding X-ray against the same text;
only strictly higher SPOOK scores count as wins.

\paragraph{Biomedical Gallery Retrieval.}
For text-to-image retrieval, each report or caption
queries the 500 real X-rays plus its own single SPOOK.
These are independent insertion trials, each containing
501 images, with ties favoring real images.
This evaluation reverses the direction of the
image-to-text ASR in Table~2 of the main paper.
The downstream illustration follows one such
query-specific trial.

\paragraph{Corpus-Scale ImageNet Retrieval.}
We reuse the same $1000$ CLIP ViT-B/16 SPOOKs from
the main ImageNet run, one per class, without further
optimization.
All $1000$ SPOOKs are inserted simultaneously into
each gallery: one contains 50,000 validation images,
and the other contains 1,281,167 training images.
SPOOKs constitute 1.96\% and 0.078\% of the augmented
galleries, respectively.
The same $1000$ class-caption queries search both
galleries.

Retrieval compares each query with every gallery image
using cosine similarity.
Saved SPOOK images are encoded again using the
evaluation model and its standard preprocessing.
Ties favor real images.
Target Top-$k$ is the percentage of queries retrieving
their designated class SPOOK among the first $k$ images.
These text-to-image rates are distinct from the main
target-bank ASR.
The two galleries come from different ImageNet splits
and provide two corpus-scale settings rather than
nested subsets.

\paragraph{Downstream Verification.}
Figure~5 of the main paper follows a representative
Open-i retrieval into downstream verification.
BiomedCLIP retrieves from 500 real X-rays and one
inserted SPOOK using the original report.
Under a top-1 evidence policy,
Qwen3-VL-8B-Instruct~[1] receives
only the retrieved SPOOK and the displayed user request.
The real X-ray, report, scores, and SPOOK label
are not supplied to the downstream model.
The report used for retrieval is separate from the
downstream request referring to pleural effusion.

Generation is deterministic.
A separate instruction asks for one short sentence
of at most 35 words.
Different request wordings were explored with the
same image; the figure preserves the full displayed
request and its first generated response.
The case illustrates how evidence selection affects
the image available for downstream verification.

\paragraph{Runtime Accounting.}
Targets are optimized in parallel batches on a single
NVIDIA H200.
Reported per-target and per-query times divide the
complete run time by the number of targets and candidate
evaluations, respectively.
They include setup and evaluation overhead and describe
average costs under batching.
Further measurements appear in
Appendix~\ref{app:runtime-analysis}.

\begin{algorithm}[t]
\caption{Targeted Sparse Embedding Spooking}
\label{alg:sparse-spoof}
\begin{algorithmic}[1]
\STATE \textbf{Input:} image encoder $f$, target bank $\mathcal{T}=\{t_1,\ldots,t_N\}$, target index $i$, initial canvas $x_0$, query budget $Q$, objective $J(x,t_i)$
\STATE \textbf{Output:} SPOOK image $x$, accepted edit count $a$, target rank $r_i$
\STATE $x \leftarrow x_0$
\STATE $a \leftarrow 0$
\STATE $\mathcal{M} \leftarrow \emptyset$ \hfill edited pixel-channel set
\STATE $s \leftarrow J(x,t_i)$
\FOR{$q=1$ to $Q$}
    \STATE sample an unedited pixel-channel $m \notin \mathcal{M}$
    \STATE sample a replacement value $v \sim \mathcal{U}(0,1)$
    \STATE $x' \leftarrow x$
    \STATE $x'_m \leftarrow v$
    \STATE $s' \leftarrow J(x',t_i)$
    \IF{$s' > s$}
        \STATE $x \leftarrow x'$
        \STATE $s \leftarrow s'$
        \STATE $\mathcal{M} \leftarrow \mathcal{M} \cup \{m\}$
        \STATE $a \leftarrow a+1$
    \ENDIF
\ENDFOR
\STATE compute $z_j=\cos(f(x),t_j)$ for all $t_j\in\mathcal{T}$
\STATE $r_i \leftarrow 1+\sum_{j\neq i}\mathbf{1}[z_j>z_i]$
\STATE \textbf{return} $x$, $a$, $r_i$
\end{algorithmic}
\end{algorithm}

\subsection{Qualitative SPOOK Examples}
\label{sec:appendix_qualitative_examples}

Figures~\ref{fig:appendix_clip_imagenet_examples}--\ref{fig:appendix_dinov3_imagenet_examples} show representative qualitative examples for the main experimental settings. Each row compares a natural reference image, the white initialization canvas, and the optimized SPOOK image for the same target. The score below each image is the target cosine similarity used by the corresponding model and dataset condition.

These examples highlight the central phenomenon studied in the paper: visually sparse, non-natural images can nevertheless achieve high target alignment under the embedding space of the evaluated model. The qualitative examples also emphasize that the optimized images do not resemble natural target instances to a human observer, supporting the interpretation that SPOOK images exploit model-specific sparse directions rather than human-semantic visual content.

\begin{table*}[t]
\centering
\caption{
\textbf{Runtime for the three main SPOOK settings.}
Each run optimizes $1000$ targets with a budget of $50{,}000$ candidate pixel-channel queries per target on a single NVIDIA H200 GPU.
}
\label{tab:runtime-analysis}
\setlength{\tabcolsep}{10pt}
\renewcommand{\arraystretch}{1.12}
\begin{tabular}{llcc}
\toprule
Dataset
& Model
& Total runtime
& Runtime / query / target \\
\midrule
ImageNet
& OpenCLIP B/16
& 2.36 h
& 0.1699 ms \\

Flickr30k
& OpenCLIP B/16
& 2.29 h
& 0.1649 ms \\

ImageNet
& DINOv3 B/16
& 3.16 h
& 0.2275 ms \\
\bottomrule
\end{tabular}

\vspace{1mm}
\begin{flushleft}
\footnotesize
Total runtime is measured for the full $1000$-target run.
Runtime / query / target is computed as total runtime divided by $50{,}000\times1000$ candidate evaluations.
\end{flushleft}
\end{table*}

\subsection{Supplementary Analysis}

\subsubsection{Interpolation paths reveal distinct sparse and natural routes}
\label{sec:interpolation_appendix}

\begin{figure*}[t]
    \centering
    \includegraphics[width=.86\linewidth]{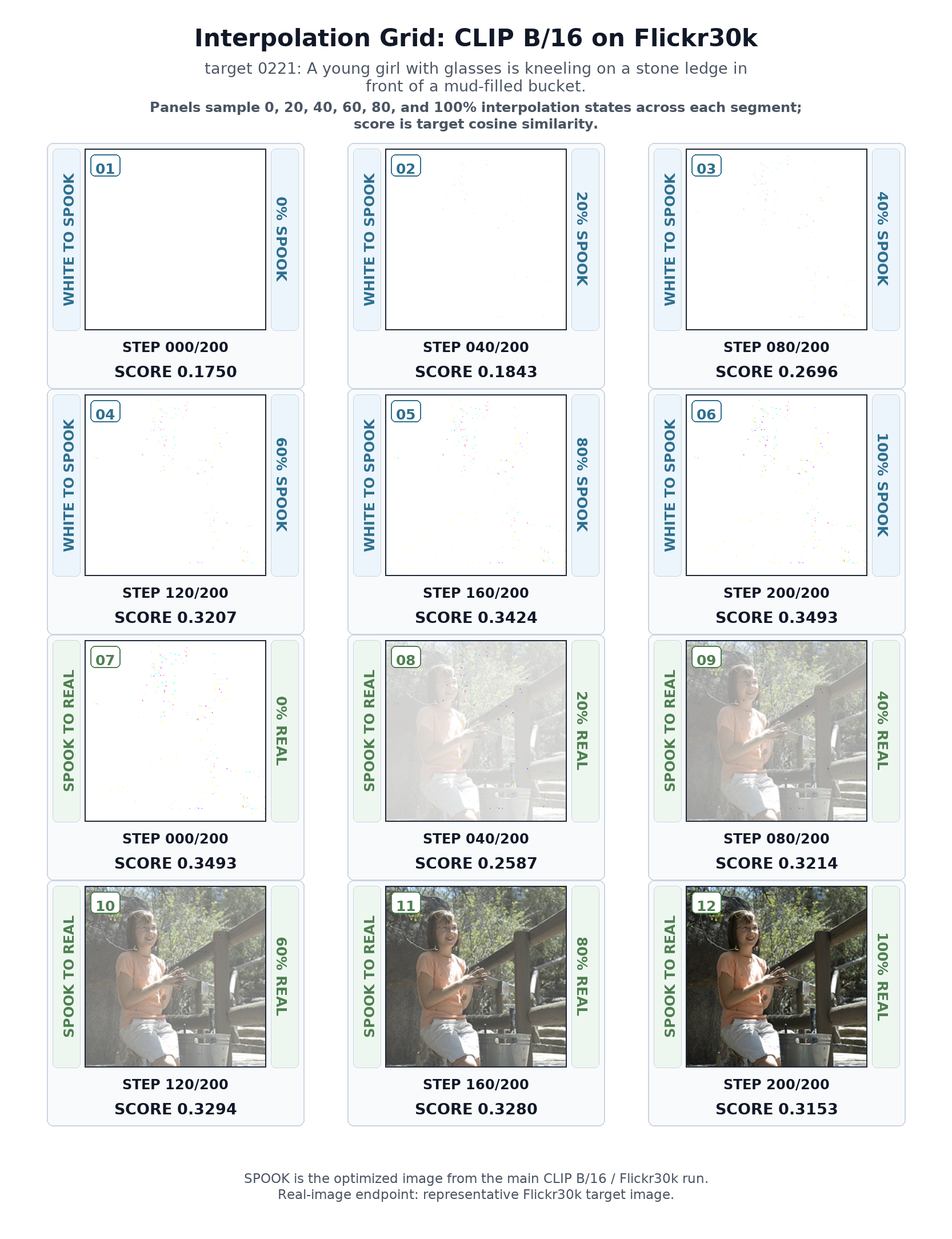}
    \caption{
    \textbf{Representative interpolation grid for CLIP/Flickr30k.}
    The target is a Flickr30k caption, and scores are CLIP image--text cosine similarities.
    The first two rows interpolate from the white canvas to the optimized SPOOK image; the last two rows interpolate from the SPOOK image to the real target image.
    The SPOOK endpoint obtains a high target score while remaining visually sparse, but mixing it with the real image produces a lower-scoring intermediate state before the natural image route recovers target similarity.
    }
    \label{fig:supp_interpolation_grid_flickr}
\end{figure*}

\begin{figure*}[t]
    \centering
    \includegraphics[width=.86\linewidth]{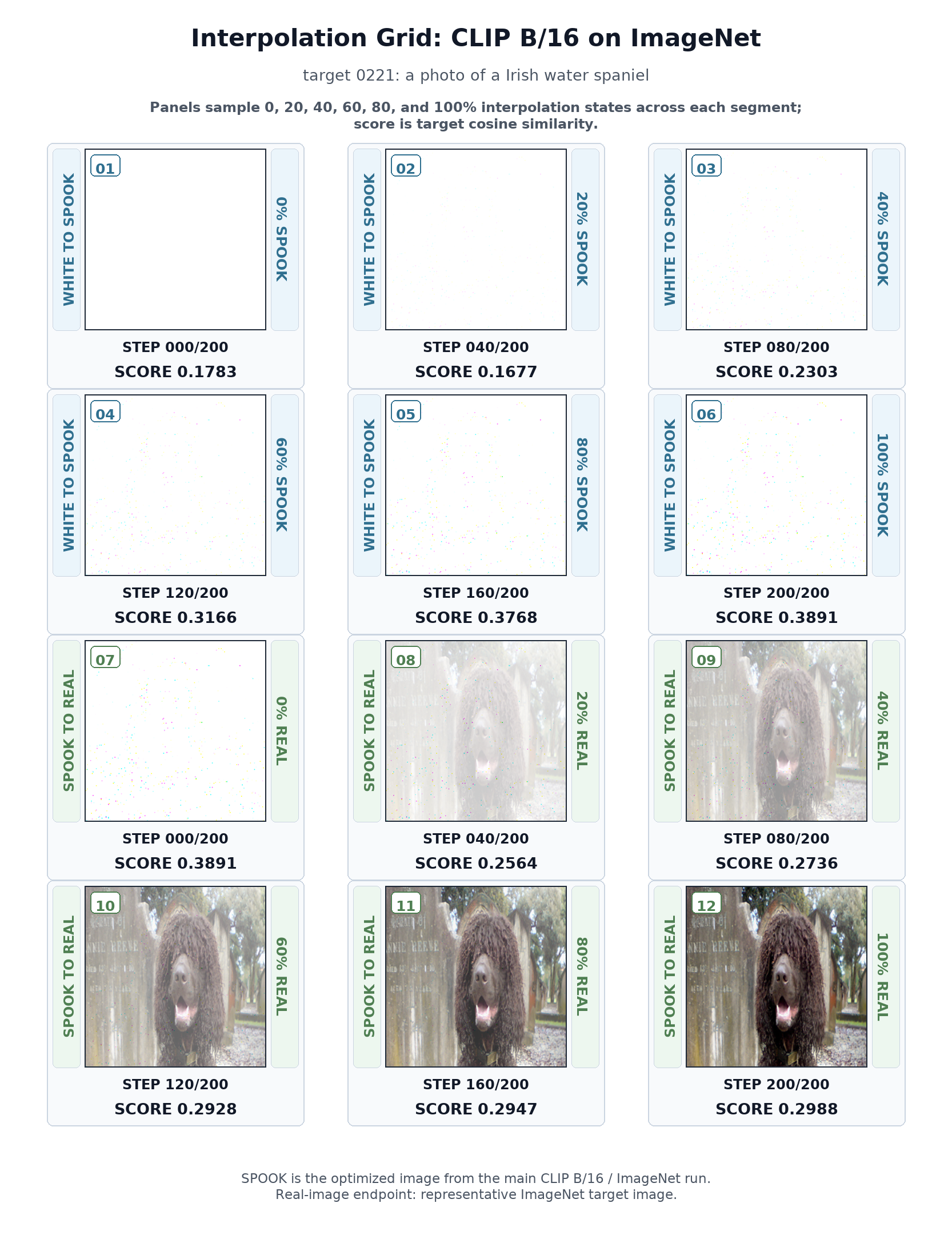}
    \caption{
    \textbf{Representative interpolation grid for CLIP/ImageNet.}
    The target is the prompt ``a photo of a Irish water spaniel,'' and scores are CLIP image--text cosine similarities.
    Sparse edits from the white canvas produce a high-scoring SPOOK image without recognizable target content.
    When the SPOOK image is interpolated toward the natural ImageNet image, the target score drops sharply before recovering near the real-image endpoint, illustrating a distinct sparse route to target alignment.
    }
    \label{fig:supp_interpolation_grid_clip_imagenet}
\end{figure*}

\begin{figure*}[t]
    \centering
    \includegraphics[width=.86\linewidth]{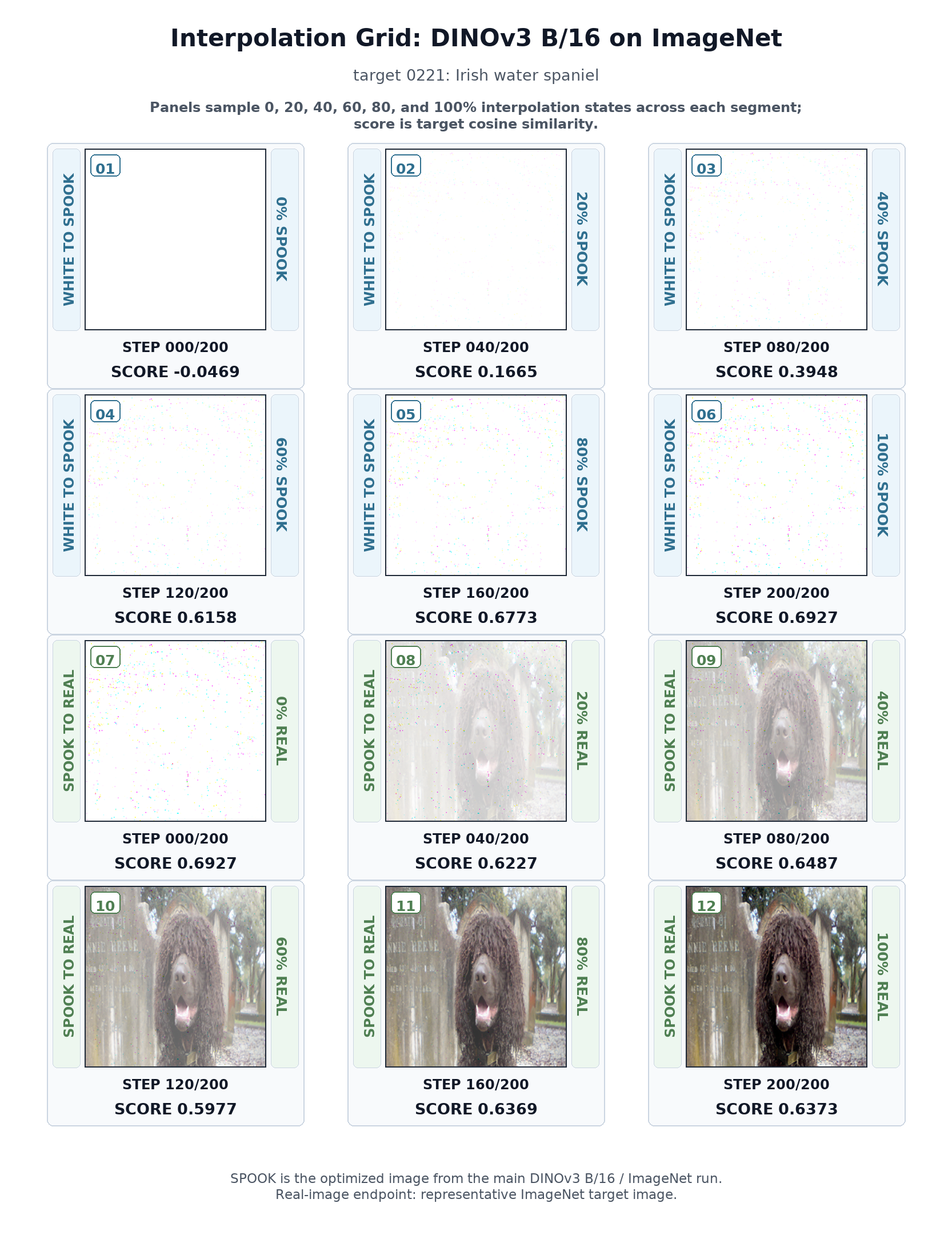}
    \caption{
    \textbf{Representative interpolation grid for DINOv3/ImageNet.}
    The target is the ImageNet class ``Irish water spaniel,'' and scores are DINOv3 image--prototype cosine similarities.
    As in the CLIP setting, sparse edits on the white canvas produce a high-scoring SPOOK image despite the absence of recognizable target content.
    Interpolating from SPOOK to the natural image produces a non-monotonic score path, indicating that the sparse and natural inputs reach the class prototype through different pixel-space routes.
    }
    \label{fig:supp_interpolation_grid_dino}
\end{figure*}

\begin{figure*}[t]
    \centering
    \includegraphics[width=\linewidth,height=0.8\textheight,keepaspectratio]{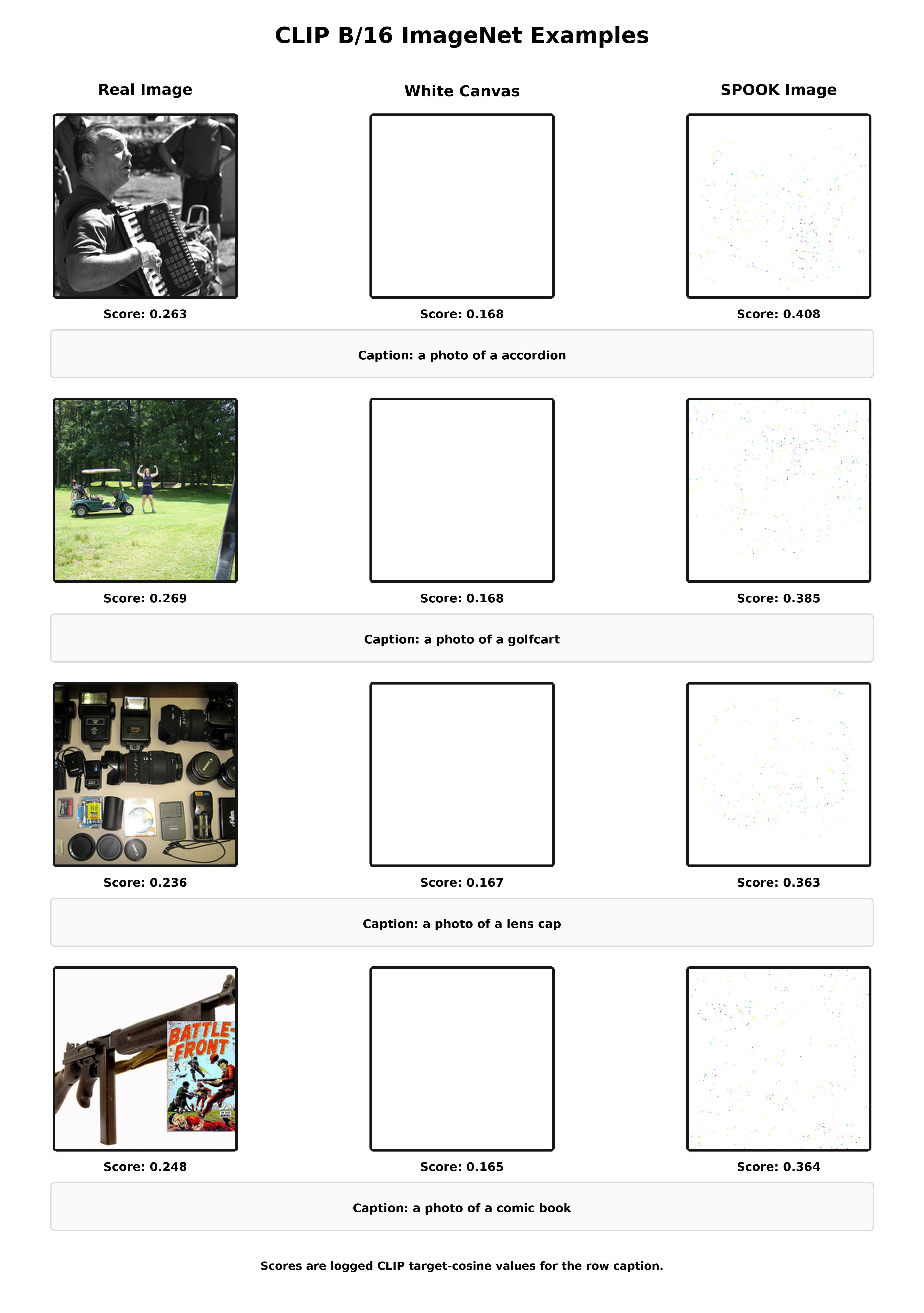}
    \caption{
    Qualitative CLIP/ImageNet examples. Each row shows a natural ImageNet image, the white canvas initialization, and the optimized SPOOK image for the same target caption. Scores are CLIP image--text cosine similarities to the row caption.
    }
    \label{fig:appendix_clip_imagenet_examples}
\end{figure*}

\begin{figure*}[t]
    \centering
    \includegraphics[width=\linewidth,height=0.8\textheight,keepaspectratio]{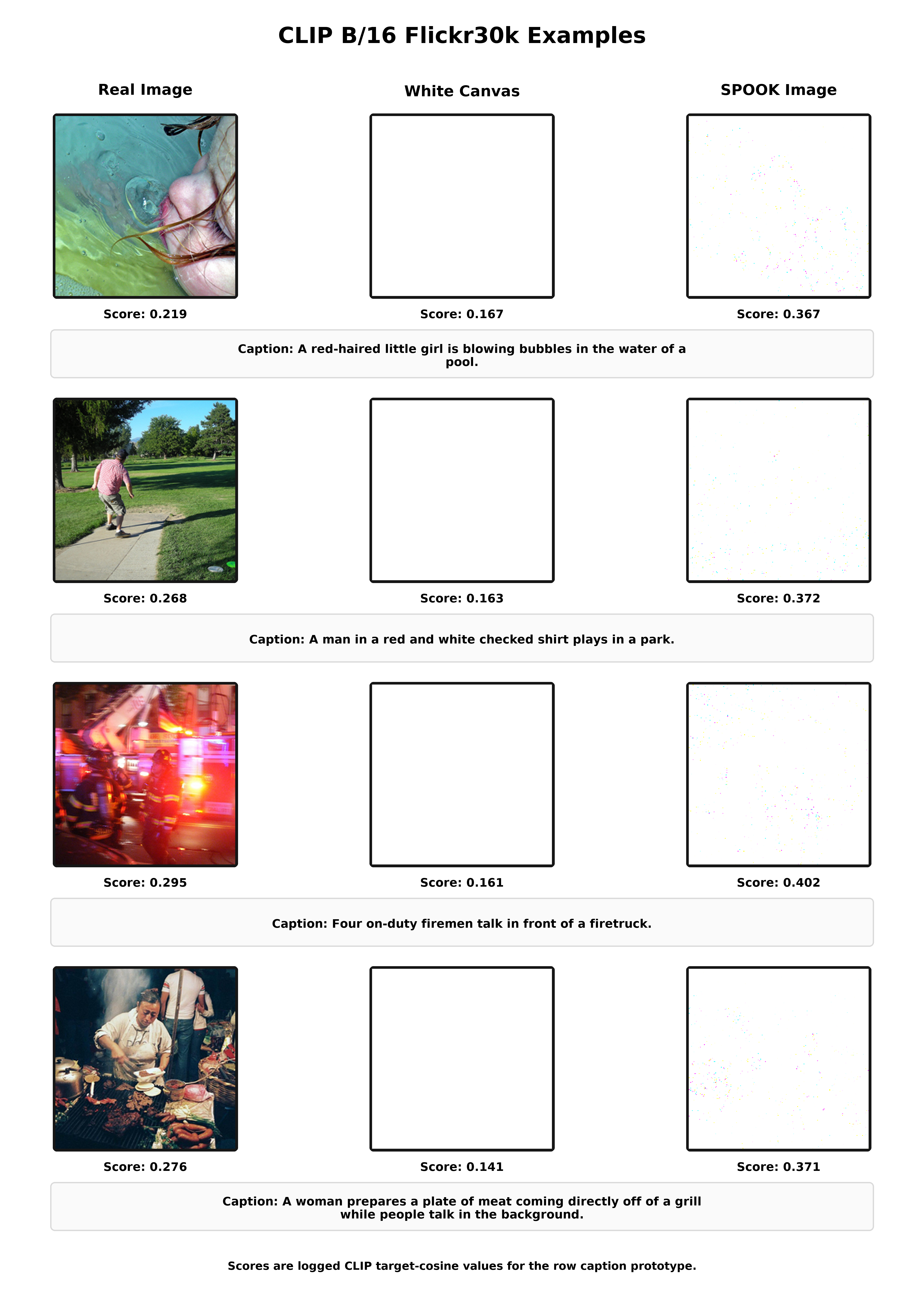}
    \caption{
    Qualitative CLIP/Flickr30k examples. Each row shows the real Flickr30k image, the white canvas initialization, and the optimized SPOOK image for the corresponding caption target. Scores are CLIP image--text cosine similarities to the row caption.
    }
    \label{fig:appendix_clip_flickr30k_examples}
\end{figure*}

\begin{figure*}[t]
    \centering
    \includegraphics[width=\linewidth,height=0.9\textheight,keepaspectratio]{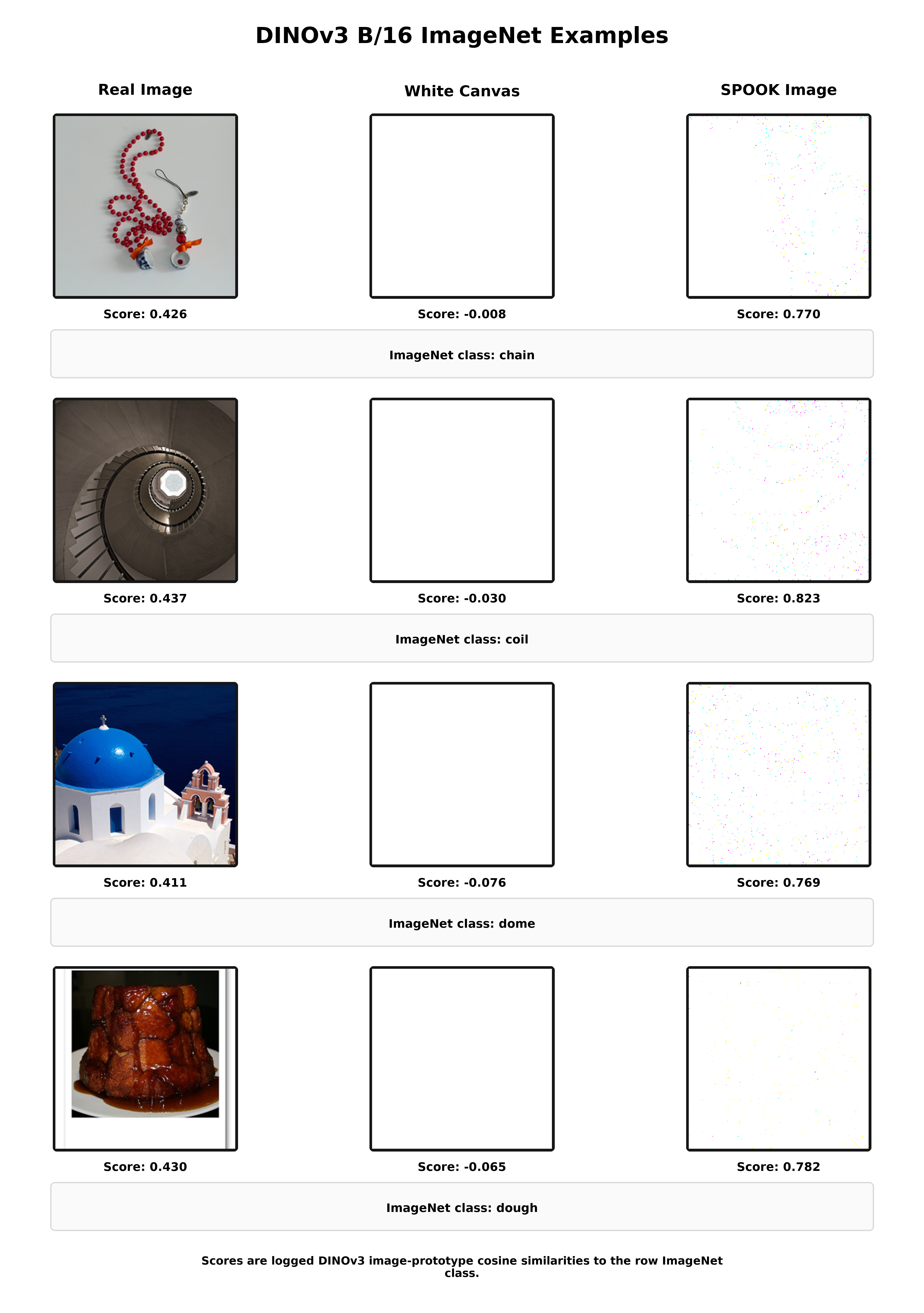}
    \caption{
    Qualitative DINOv3/ImageNet examples. Each row shows a natural ImageNet image, the white canvas initialization, and the optimized SPOOK image for the same ImageNet class target. Scores are DINOv3 image--prototype cosine similarities to the row ImageNet class.
    }
    \label{fig:appendix_dinov3_imagenet_examples}
\end{figure*}

To further probe whether SPOOK images behave like weak or partial versions of natural images, we visualize smooth image-space interpolation paths among three endpoints: the white canvas, the optimized SPOOK image, and a corresponding real target image.
Figure~4 of the main paper summarizes target cosine scores along these paths for the three main model/dataset settings, and Figures~\ref{fig:supp_interpolation_grid_flickr}--\ref{fig:supp_interpolation_grid_dino} show representative interpolation grids.

The score profiles reveal a consistent qualitative pattern.
Along the white-to-SPOOK path, target similarity generally increases as sparse edits are introduced.
However, along the SPOOK-to-real path, similarity often drops sharply when the sparse SPOOK image is mixed with the natural image, before recovering as the real image dominates.
This non-monotonic behavior suggests that SPOOK and natural images do not simply lie on the same visual route to the target representation.
Instead, the encoder can assign high target similarity through two distinct pixel-space mechanisms: a sparse off-manifold route and a natural-image route.
The direct white-to-real path further supports this interpretation, showing that the sparse SPOOK endpoint is not merely an intermediate point along ordinary interpolation from blank canvas to natural image.

These visualizations complement the main interpolation analysis in Figure~4 of the main paper.
They do not imply that SPOOK images contain hidden human-recognizable semantics; rather, they illustrate that target alignment in embedding space can be reached through visually different and geometrically distinct input paths.

\subsubsection{Cross-model transfer: SPOOKs are model-specific}
\label{sec:transfer}

We next test whether SPOOKs optimized for one encoder transfer to another. Transfer is weak across model families. CLIP ViT-B/16 SPOOKs achieve $93.9\%$ ASR@1 when evaluated on the same CLIP model, but only $0.6\%$ ASR@1 and $2.1\%$ ASR@5 when evaluated on DINOv3 ViT-B/16. Conversely, DINOv3 ViT-B/16 SPOOKs achieve $85.5\%$ ASR@1 on their source model, but only $0.2\%$ ASR@1 and $0.8\%$ ASR@5 when evaluated on CLIP ViT-B/16.

Transfer is also weak even within the CLIP family. CLIP ViT-B/16 SPOOKs evaluated on CLIP ViT-B/32 achieve only $1.1\%$ ASR@1 and $3.9\%$ ASR@5, while CLIP ViT-B/32 SPOOKs evaluated on CLIP ViT-B/16 achieve $0.4\%$ ASR@1 and $2.0\%$ ASR@5. These results indicate that SPOOK images are not model-independent semantic depictions of the target. They are model-specific off-manifold embedding triggers.

The lack of transfer clarifies the threat model. SPOOK images need not be universal semantic images to matter: many retrieval, filtering, indexing, and nearest-neighbor systems deploy a fixed encoder. For such systems, a model-specific trigger is sufficient to induce targeted retrieval failures.

\subsubsection{Robustness to Simple Image Transformations}
\label{sec:spook-robustness}

Sparse SPOOK images are highly effective in the exact evaluation setting, but they need not be invariant to standard image transformations. To check this, we run a minimal robustness test on the three main-table conditions. For each condition, we evaluate a fixed random subset of 64 targets, while preserving the original 1000-target retrieval bank for rank and ASR computation. We apply JPEG compression, center cropping, Gaussian blur, and down-up resizing to the final SPOOK image before re-encoding it.

Table~\ref{tab:spook_robustness} shows that SPOOK images are substantially brittle to these transformations. Even mild JPEG compression or small center crops sharply reduce ASR@1 and ASR@5 across CLIP and DINOv3. This supports treating transformation robustness as a separate property from targeted retrieval success in the untransformed setting.

\subsubsection{Runtime analysis}
\label{app:runtime-analysis}

SPOOK is query-based, but the implementation is efficient because targets are optimized in parallel batches. Table~\ref{tab:runtime-analysis} reports wall-clock runtime for the three main settings on a single NVIDIA H200 GPU. Each run optimizes $1000$ targets with $50{,}000$ candidate pixel-channel queries per target, and completes in roughly $2$--$3$ GPU-hours.

Despite the large nominal query budget, SPOOK remains practical at this scale: the main $1000$-target experiments complete within a few GPU-hours using a batched implementation.

\subsection{Discussion}
\label{sec:discussion}

\paragraph{Alignment and Visual Evidence.}
SPOOK separates an encoder's preference for a target
from recognizable support for that target in the image.
The occurrence across language-aligned, image-only,
and biomedical representations broadens the scope
of this mismatch across the settings examined.
High target similarity can therefore be a poor basis
for deciding whether a retrieved image supplies
usable visual evidence.

\paragraph{Role of the Construction.}
The controls connect successful alignment to
target-specific optimization and the initial canvas.
Matched random edits do not reproduce the effect,
while uniform backgrounds favor alignment over
dense random initialization.
Sparse blank canvases thus provide a controlled
way to examine target alignment in the absence
of recognizable content.
The interpolation profiles further show that adding
real visual content can reduce target similarity,
even when both endpoints are highly aligned.

\paragraph{Consequences for Evidence Selection.}
The corpus-scale results connect this representation
behavior to retrieval under competition with real images.
With image-insertion access and target-score queries,
SPOOK can occupy retrieval slots without changing
the encoder or existing corpus.
When a pipeline passes only its top-ranked image
downstream, this selection determines the available
visual evidence.
The biomedical example illustrates the consequence:
the model correctly identifies an unusable reference,
yet cannot inspect the real images excluded by retrieval.
A downstream validity check can detect the problem
while leaving the verification request unresolved.

\subsection{Limitations}

The observed alignment is model-specific, with weak
transfer across encoders and sensitivity to compression,
cropping, blur, and resizing.
Optimization also requires target-score access and
a substantial query budget despite the small number
of accepted edits.
The resulting images are visibly artificial, limiting
their stealth.
These conditions place the demonstrated risk in
digital pipelines that admit such images and preserve
them through retrieval.

\begin{figure*}[t]
\centering
\includegraphics[width=0.32\linewidth]{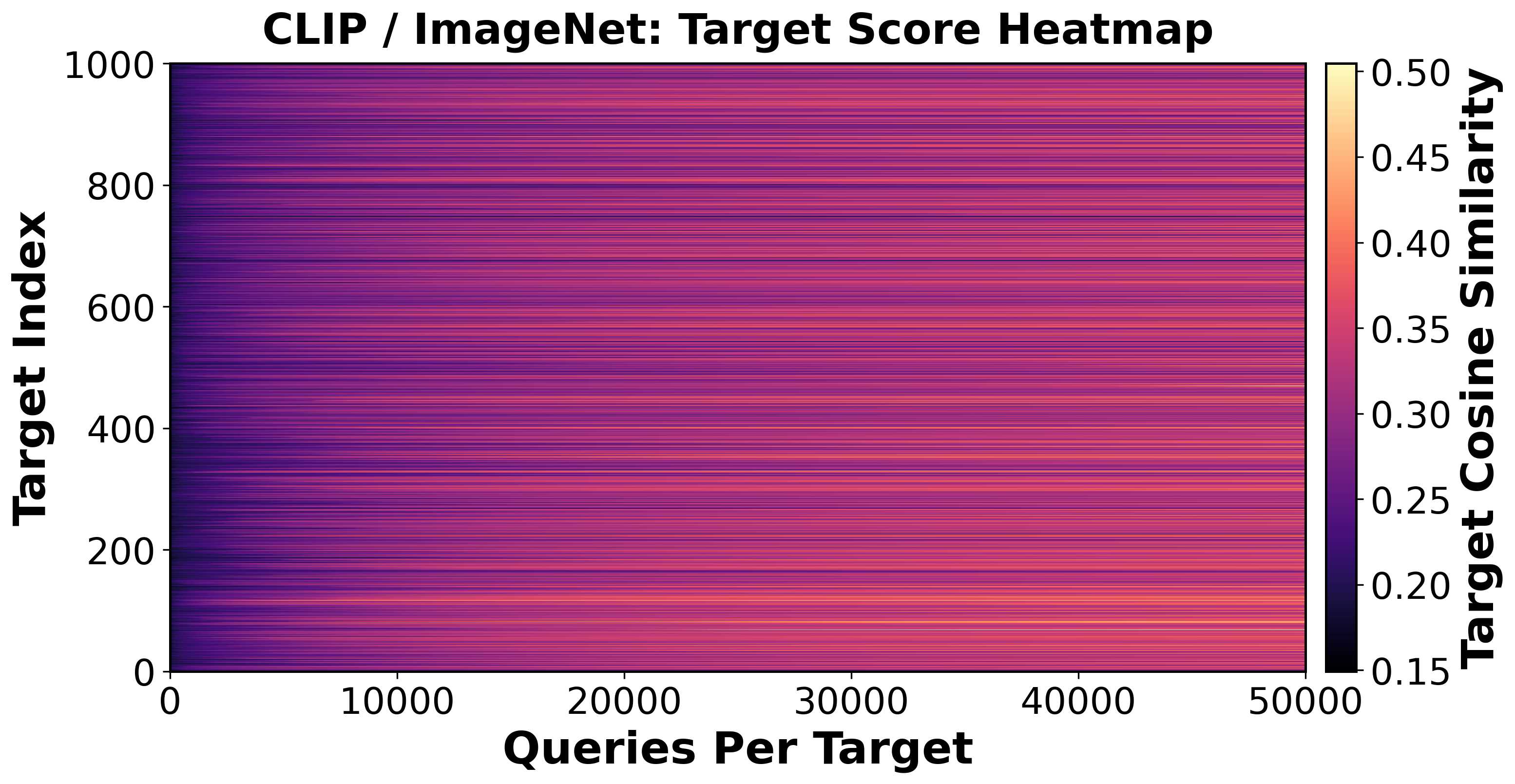}\hfill
\includegraphics[width=0.32\linewidth]{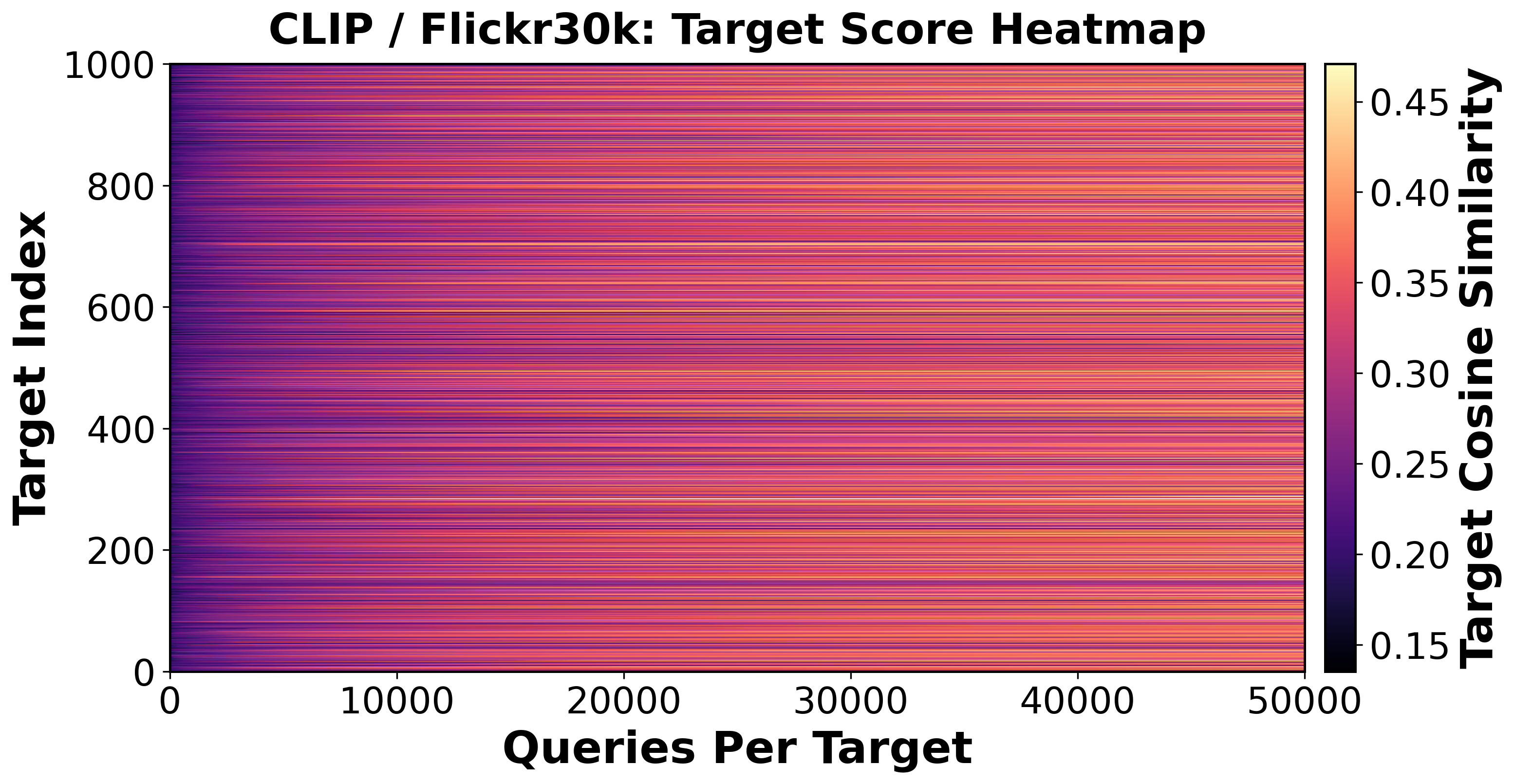}\hfill
\includegraphics[width=0.32\linewidth]{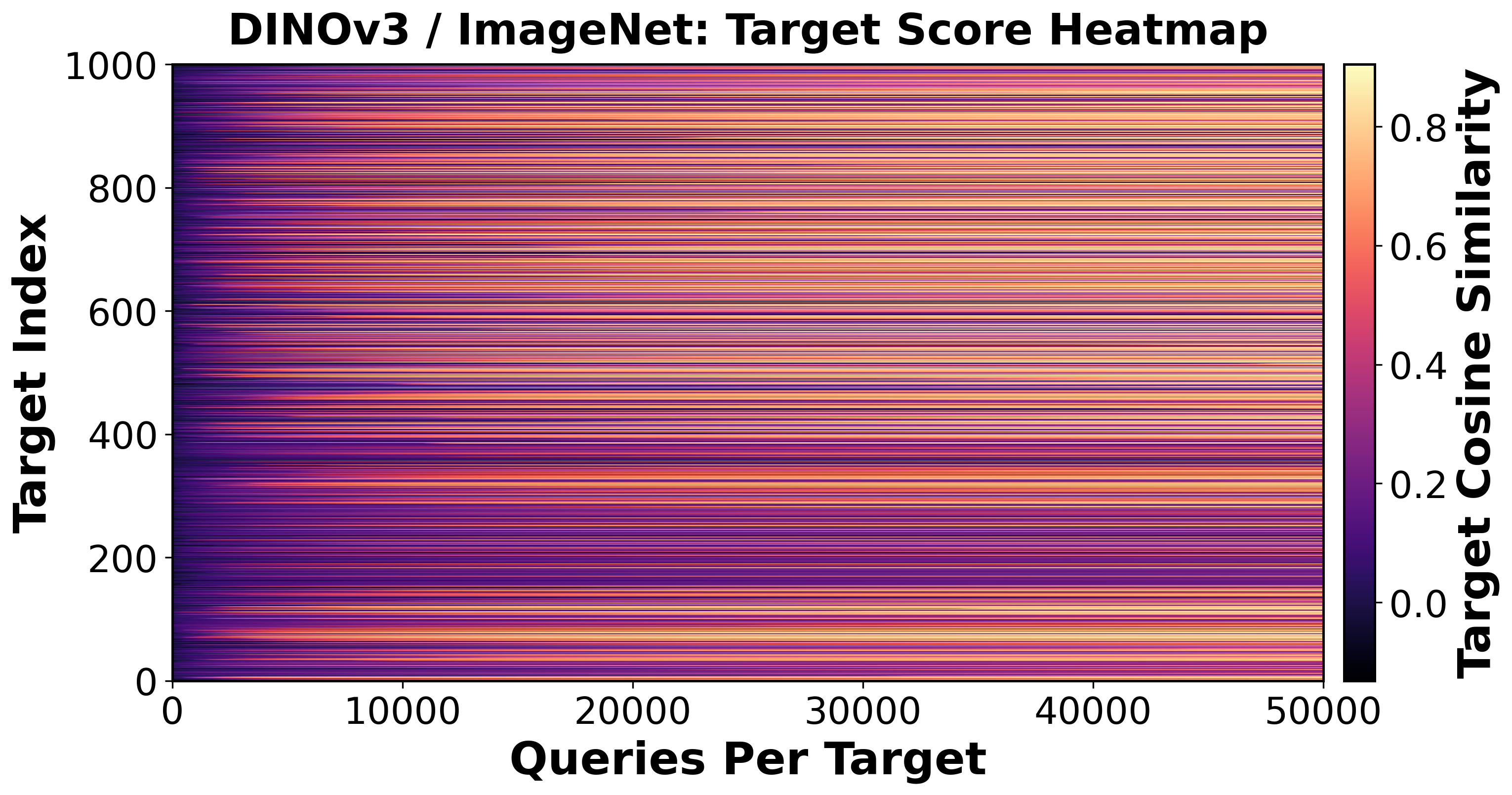}

\par\vspace{2mm}

\includegraphics[width=0.32\linewidth]{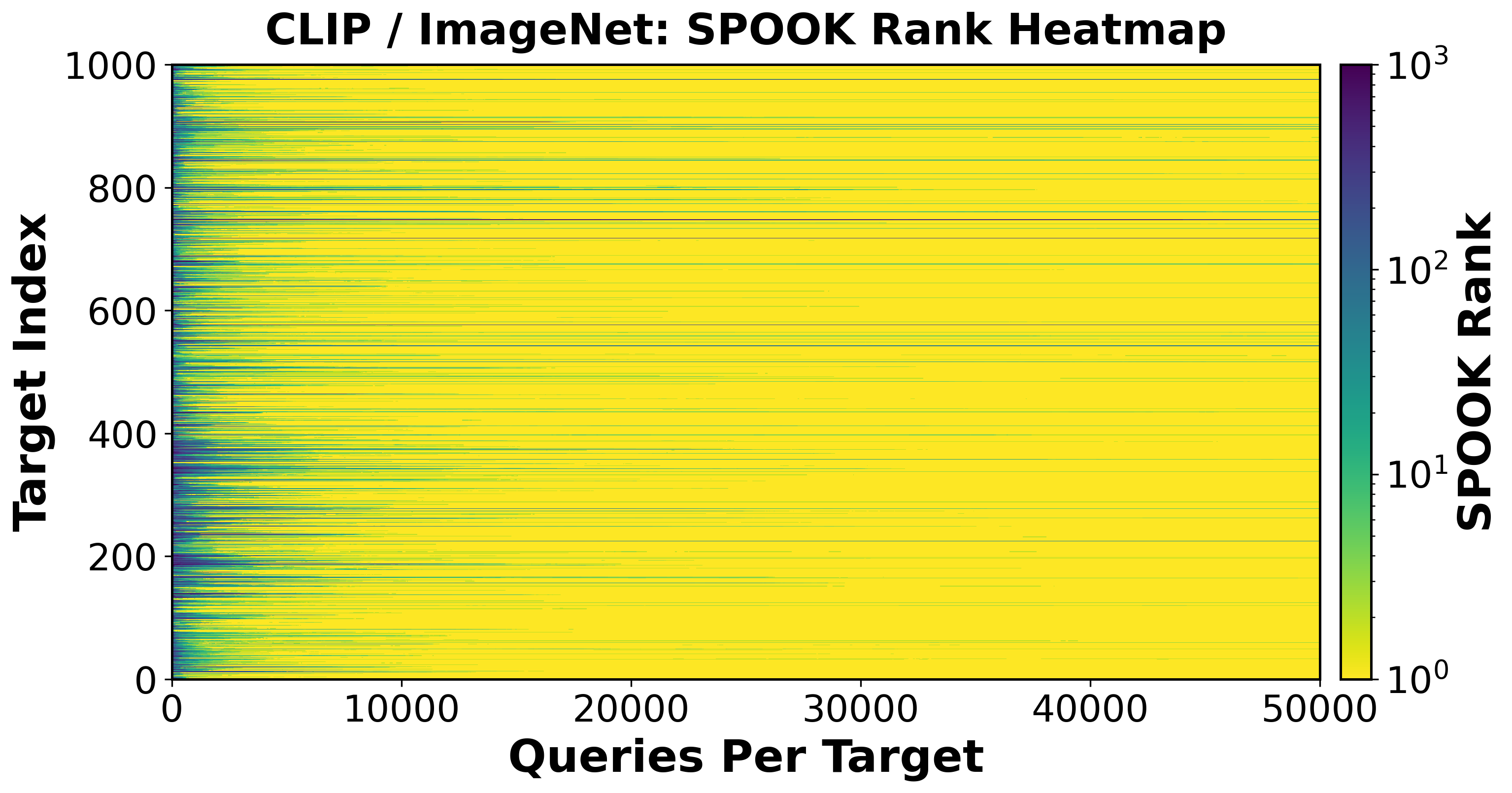}\hfill
\includegraphics[width=0.32\linewidth]{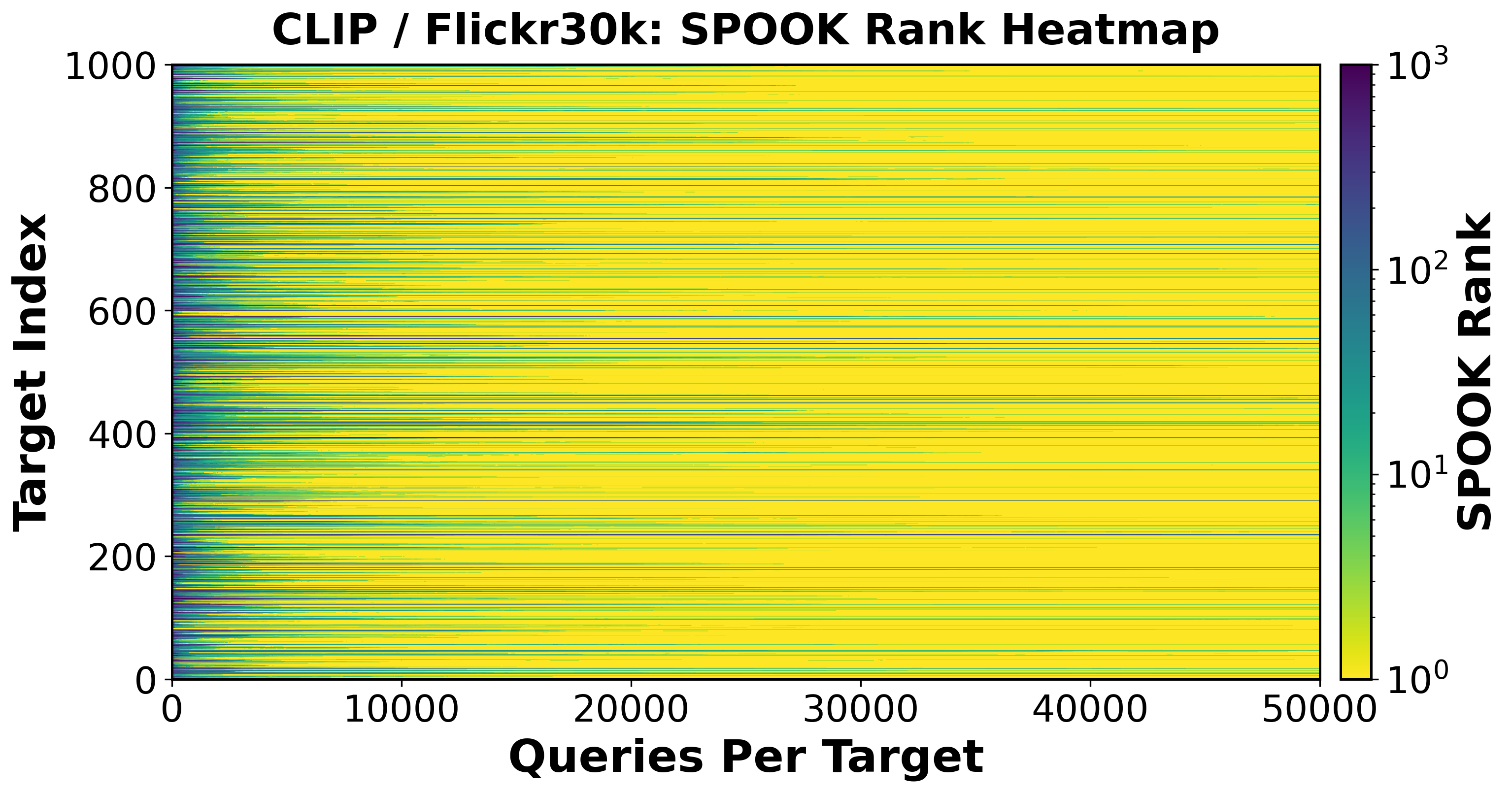}\hfill
\includegraphics[width=0.32\linewidth]{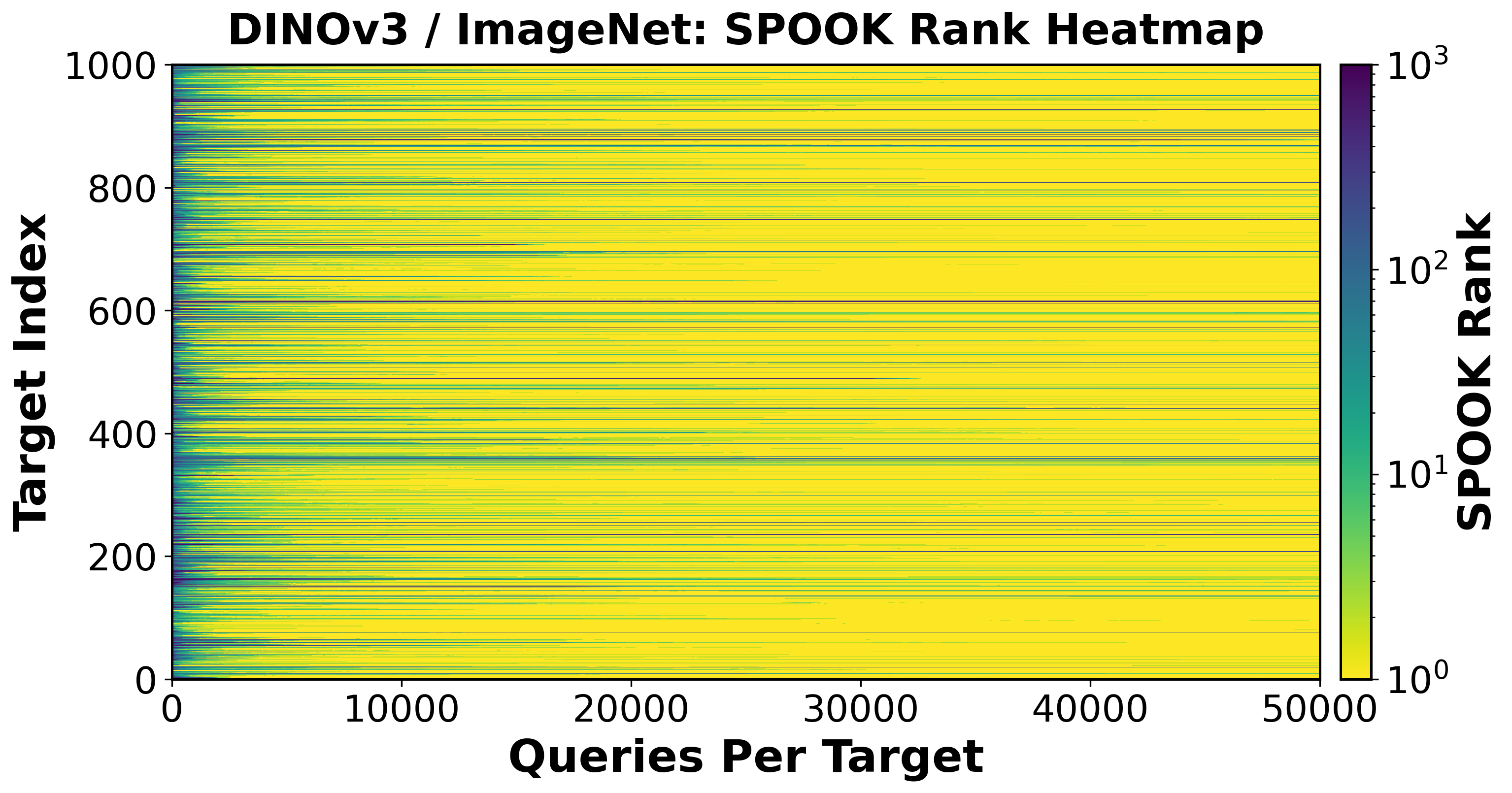}

\par\vspace{2mm}

\includegraphics[width=0.32\linewidth]{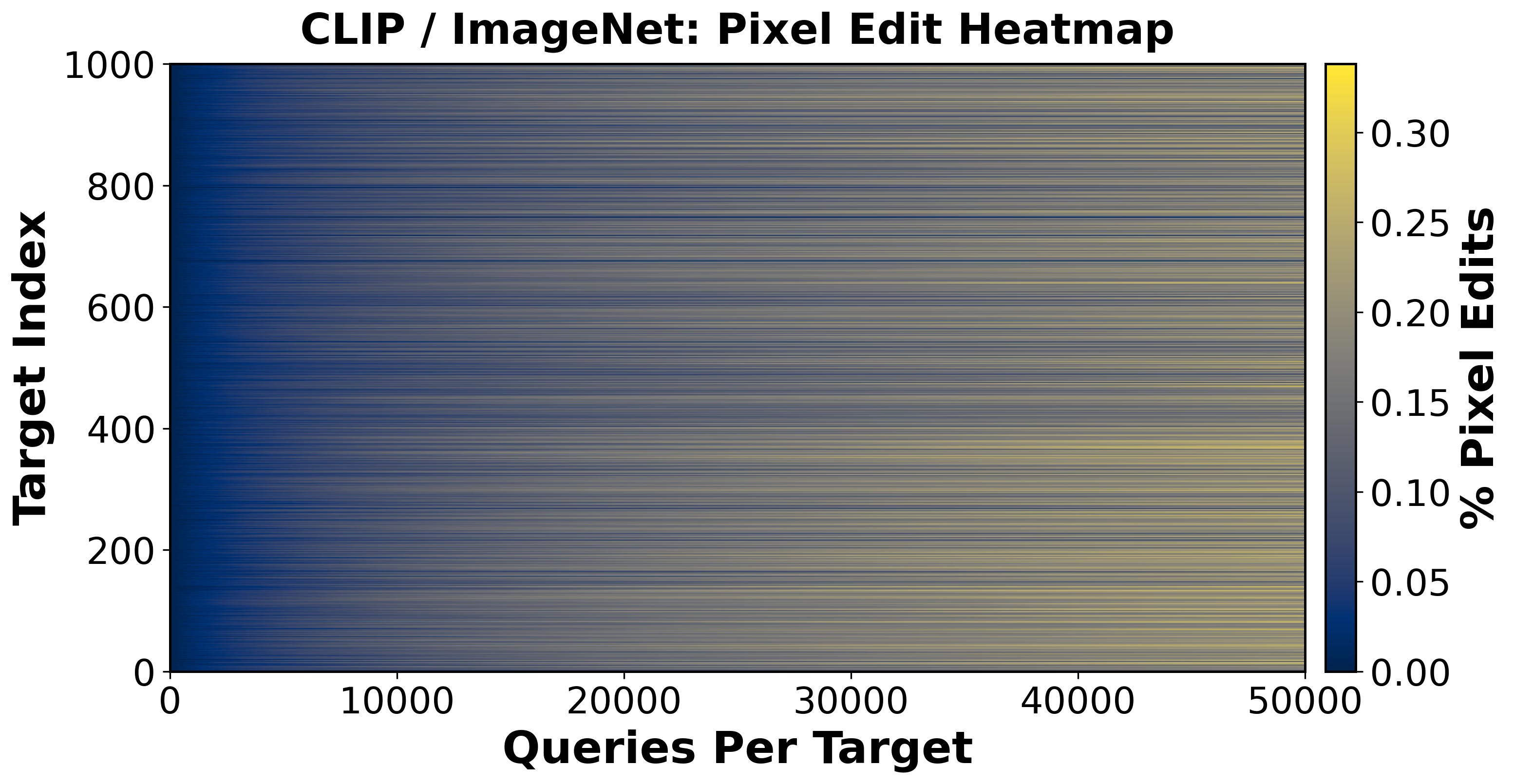}\hfill
\includegraphics[width=0.32\linewidth]{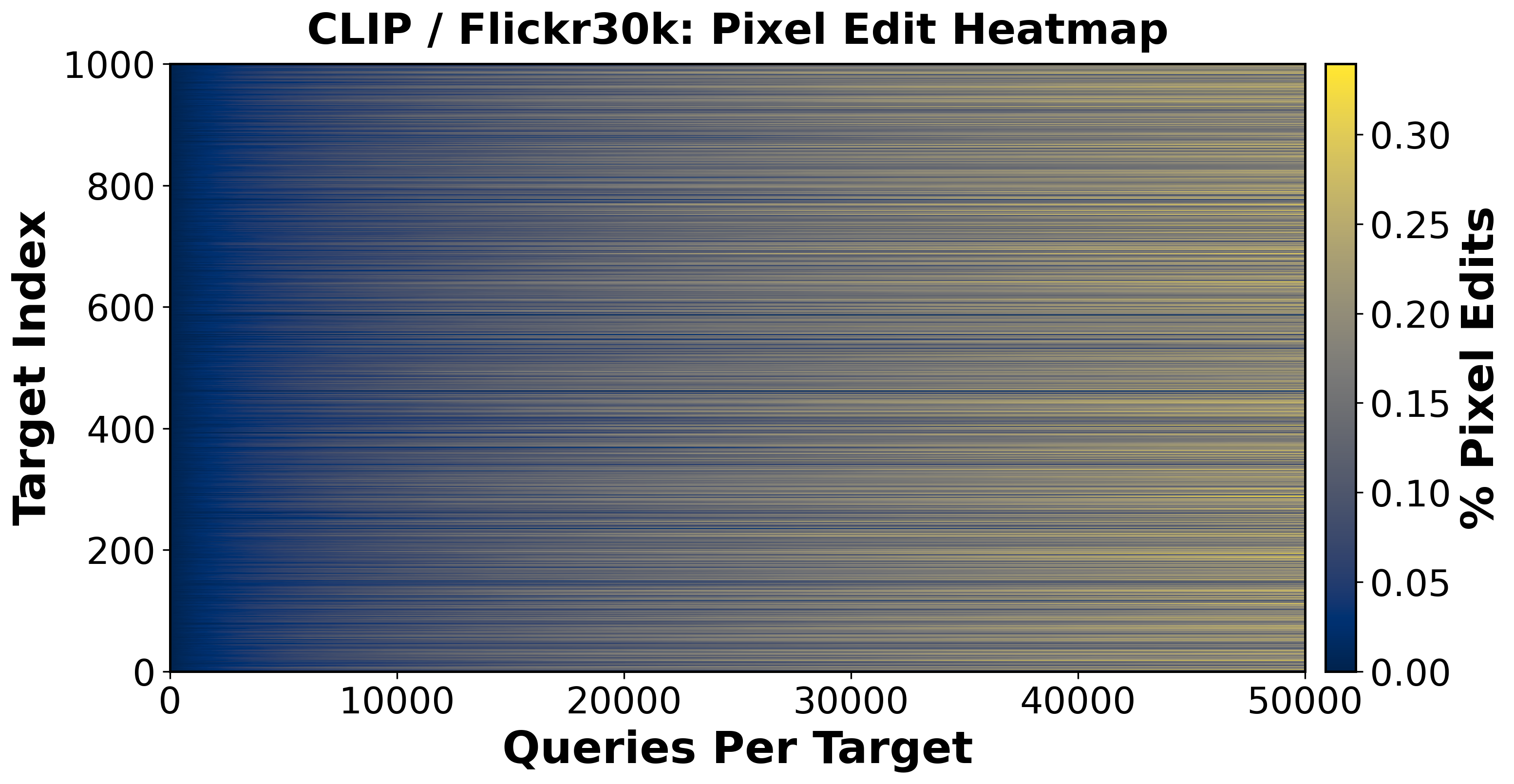}\hfill
\includegraphics[width=0.32\linewidth]{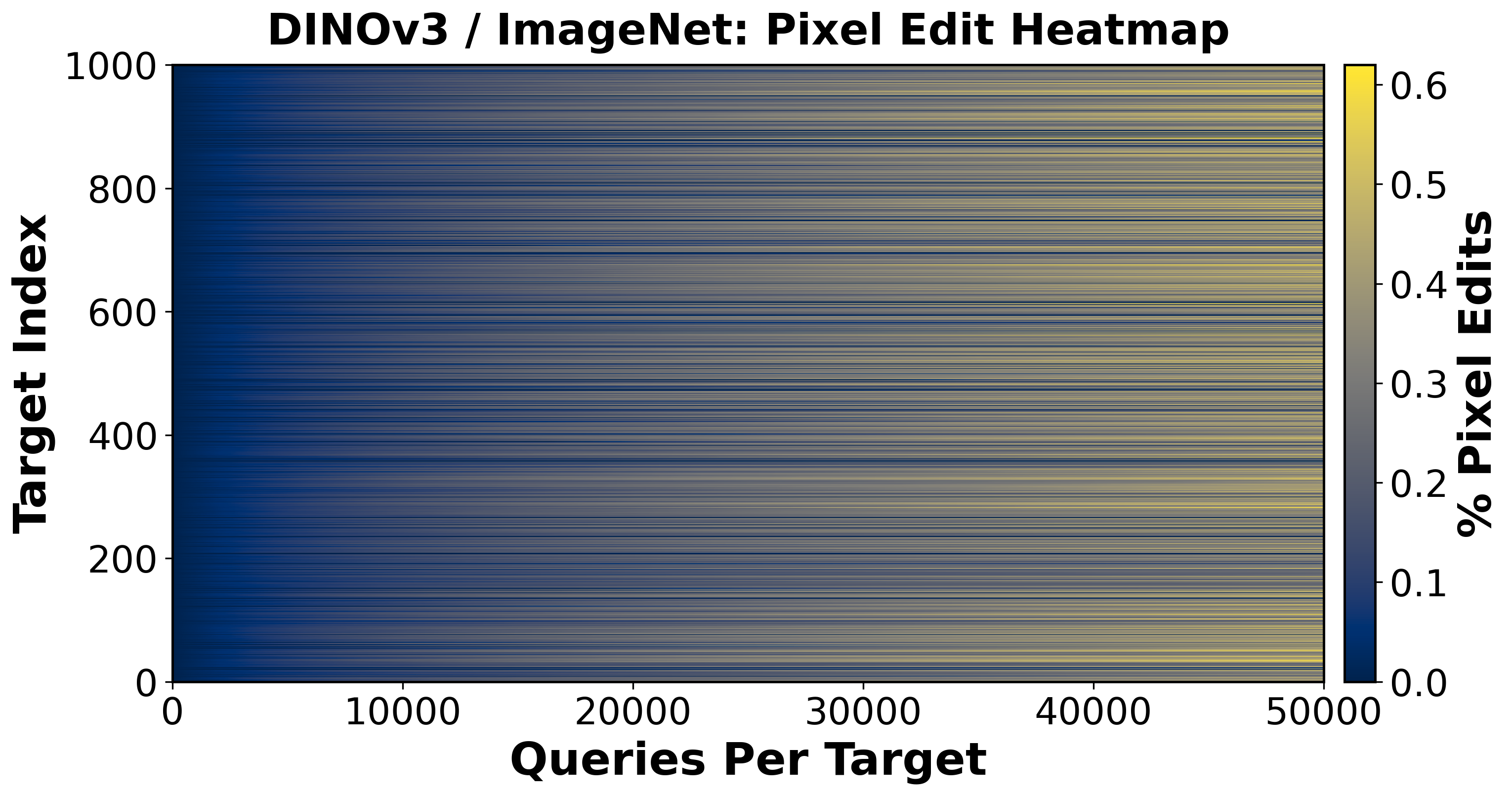}

\caption{\textbf{Per-Target SPOOK Optimization Dynamics.}
Columns show CLIP/ImageNet, CLIP/Flickr30k, and DINOv3/ImageNet.
Rows show target cosine similarity, target retrieval rank,
and cumulative pixel-channel edit percentage.
Each heatmap tracks 1,000 targets as the query budget increases.
These results reveal variation across individual targets
behind the aggregate results in Table~1 of the main paper
and Figure~3 of the main paper: similarity generally
increases and retrieval ranks improve while edits remain sparse.}
\label{fig:supp-spatial-diagnostics}
\end{figure*}

\begin{table*}[t]
\centering
\caption{
\textbf{Random sparse images do not explain SPOOK retrieval.}
For each target, the random sparse control matches SPOOK's edit count but chooses pixel-channel locations and values randomly.
All results use $1000$ targets and ViT-B/16 encoders.
}
\label{tab:random-sparse-control}
\setlength{\tabcolsep}{5pt}
\renewcommand{\arraystretch}{1.12}
\begin{tabular}{lllcccc}
\toprule
Dataset
& Model
& Method
& SPOOK cosine $\uparrow$
& ASR@1 $\uparrow$
& ASR@5 $\uparrow$
& Rank $\downarrow$ \\
\midrule
ImageNet
& CLIP
& SPOOK greedy
& 0.3314 / 0.3349
& 94.1\%
& 98.5\%
& 2.1 / 1.0 \\

ImageNet
& CLIP
& Random sparse
& 0.1384 / 0.1385
& 0.1\%
& 0.5\%
& 499.0 / 504.0 \\

\midrule
Flickr30k
& CLIP
& SPOOK greedy
& 0.3510 / 0.3549
& 87.6\%
& 95.8\%
& 2.3 / 1.0 \\

Flickr30k
& CLIP
& Random sparse
& 0.1008 / 0.0999
& 0.1\%
& 0.5\%
& 502.5 / 496.0 \\

\midrule
ImageNet
& DINOv3
& SPOOK greedy
& 0.5365 / 0.6370
& 85.8\%
& 94.9\%
& 8.5 / 1.0 \\

ImageNet
& DINOv3
& Random sparse
& -0.0196 / -0.0177
& 0.1\%
& 0.4\%
& 513.9 / 509.0 \\
\bottomrule
\end{tabular}

\vspace{1mm}
\begin{flushleft}
\footnotesize
Cosine and rank entries are reported as mean / median.
ASR values are percentages.
Ranks are computed among $1000$ targets.
For each target, the random sparse control uses the same number of edited pixel-channels as the corresponding SPOOK image, but with random locations and random values.
\end{flushleft}
\end{table*}

\begin{table*}[t]
\centering
\caption{
\textbf{Initialization strongly controls sparse spooking.}
All results use ViT-B/16 encoders, a retrieval bank of $1000$ targets, and the same target/objective settings as Table~1 of the main paper.
Only the initial canvas is changed.
}
\label{tab:init-ablation}
\setlength{\tabcolsep}{5.5pt}
\renewcommand{\arraystretch}{.95}
\begin{tabular}{llcccccc}
\toprule
& & \multicolumn{2}{c}{CLIP ImageNet}
& \multicolumn{2}{c}{CLIP Flickr30k}
& \multicolumn{2}{c}{DINOv3 ImageNet} \\
\cmidrule(lr){3-4} \cmidrule(lr){5-6} \cmidrule(lr){7-8}
Init & Metric
& Mean & Median
& Mean & Median
& Mean & Median \\
\midrule
\multirow{5}{*}{White}
& Blank cosine $\uparrow$
& 0.1763 & 0.1762
& 0.1771 & 0.1766
& -0.0355 & -0.0338 \\
& SPOOK cosine $\uparrow$
& \textbf{0.3314} & \textbf{0.3349}
& \textbf{0.3510} & \textbf{0.3549}
& \textbf{0.5365} & \textbf{0.6370} \\
& SPOOK rank $\downarrow$
& 2.1 & 1.0
& 2.3 & 1.0
& 8.5 & 1.0 \\
& Pixel edits $\downarrow$
& 0.29\% & 0.29\%
& 0.28\% & 0.28\%
& 0.61\% & 0.63\% \\
\cmidrule(lr){2-8}
& ASR@1 / ASR@5 $\uparrow$
& \multicolumn{2}{c}{94.1\% / 98.5\%}
& \multicolumn{2}{c}{87.6\% / 95.8\%}
& \multicolumn{2}{c}{85.8\% / 94.9\%} \\
\midrule
\multirow{5}{*}{Black}
& Blank cosine $\uparrow$
& 0.1890 & 0.1895
& 0.1822 & 0.1825
& -0.0147 & -0.0153 \\
& SPOOK cosine $\uparrow$
& 0.2595 & 0.2413
& 0.2549 & 0.2433
& 0.4585 & 0.5441 \\
& SPOOK rank $\downarrow$
& 66.5 & 6.0
& 125.5 & 28.0
& 14.1 & 1.0 \\
& Pixel edits $\downarrow$
& 0.11\% & 0.07\%
& 0.08\% & 0.05\%
& 0.48\% & 0.50\% \\
\cmidrule(lr){2-8}
& ASR@1 / ASR@5 $\uparrow$
& \multicolumn{2}{c}{35.3\% / 49.9\%}
& \multicolumn{2}{c}{19.6\% / 32.4\%}
& \multicolumn{2}{c}{69.4\% / 86.0\%} \\
\midrule
\multirow{5}{*}{Random}
& Blank cosine $\uparrow$
& 0.1559 & 0.1551
& 0.0986 & 0.0976
& -0.0023 & -0.0026 \\
& SPOOK cosine $\uparrow$
& 0.1654 & 0.1644
& 0.1190 & 0.1182
& 0.0029 & 0.0030 \\
& SPOOK rank $\downarrow$
& 455.3 & 441.0
& 459.4 & 451.0
& 463.1 & 443.0 \\
& Pixel edits $\downarrow$
& 0.03\% & 0.03\%
& 0.04\% & 0.04\%
& 0.04\% & 0.02\% \\
\cmidrule(lr){2-8}
& ASR@1 / ASR@5 $\uparrow$
& \multicolumn{2}{c}{0.1\% / 0.7\%}
& \multicolumn{2}{c}{0.2\% / 0.7\%}
& \multicolumn{2}{c}{0.1\% / 1.3\%} \\
\bottomrule
\end{tabular}
\end{table*}

\begin{table*}[t]
\centering
\caption{
\textbf{Objective-mode ablation for CLIP.}
All rows use CLIP ViT-B/16, white-canvas initialization, a retrieval bank of $1000$ targets, and the same query budget as Table~1 of the main paper.
Only the optimized objective is changed.
}
\label{tab:clip-objective-ablation}
\setlength{\tabcolsep}{6pt}
\renewcommand{\arraystretch}{1.15}
\begin{tabular}{llcccc}
\toprule
& & \multicolumn{2}{c}{ImageNet}
& \multicolumn{2}{c}{Flickr30k} \\
\cmidrule(lr){3-4} \cmidrule(lr){5-6}
Objective & Metric
& Mean & Median
& Mean & Median \\
\midrule
\multirow{5}{*}{Caption}
& Blank cosine $\uparrow$
& 0.1763 & 0.1762
& 0.1771 & 0.1766 \\
& SPOOK cosine $\uparrow$
& \textbf{0.3314} & \textbf{0.3349}
& \textbf{0.3510} & \textbf{0.3549} \\
& SPOOK rank $\downarrow$
& 2.1 & 1.0
& 2.3 & 1.0 \\
& Pixel edits $\downarrow$
& 0.29\% & 0.29\%
& 0.28\% & 0.28\% \\
\cmidrule(lr){2-6}
& ASR@1 / ASR@5 $\uparrow$
& \multicolumn{2}{c}{94.1\% / 98.5\%}
& \multicolumn{2}{c}{87.6\% / 95.8\%} \\
\midrule
\multirow{5}{*}{Image}
& Blank cosine $\uparrow$
& 0.3121 & 0.3060
& 0.2296 & 0.2313 \\
& SPOOK cosine $\uparrow$
& 0.6234 & 0.5923
& 0.5292 & 0.5118 \\
& SPOOK rank $\downarrow$
& 117.4 & 18.0
& 28.6 & 2.0 \\
& Pixel edits $\downarrow$
& 0.34\% & 0.28\%
& 0.36\% & 0.34\% \\
\cmidrule(lr){2-6}
& ASR@1 / ASR@5 $\uparrow$
& \multicolumn{2}{c}{27.8\% / 41.9\%}
& \multicolumn{2}{c}{41.2\% / 58.9\%} \\
\midrule
\multirow{5}{*}{Joint}
& Blank cosine $\uparrow$
& 0.2442 & 0.2408
& 0.2034 & 0.2030 \\
& SPOOK cosine $\uparrow$
& 0.5074 & 0.5275
& 0.4809 & 0.4930 \\
& SPOOK rank $\downarrow$
& 19.2 & 1.0
& 3.2 & 1.0 \\
& Pixel edits $\downarrow$
& 0.45\% & 0.46\%
& 0.48\% & 0.49\% \\
\cmidrule(lr){2-6}
& ASR@1 / ASR@5 $\uparrow$
& \multicolumn{2}{c}{74.0\% / 83.6\%}
& \multicolumn{2}{c}{86.2\% / 94.5\%} \\
\bottomrule
\end{tabular}

\vspace{1mm}
\begin{flushleft}
\footnotesize
Cosine, rank, and pixel-edit rows are reported as mean / median.
ASR values are reported once per setting.
Ranks are computed among $1000$ targets.
Pixel edits are percentages of edited pixel-channels out of $224\times224\times3$.
For caption mode, the target is a text embedding; for image mode, the target is an image embedding or prototype; for joint mode, the optimized objective combines caption and image similarity.
\end{flushleft}
\end{table*}

\begin{table*}[t]
\centering
\caption{
\textbf{Sparse SPOOKs show little cross-model transfer.}
Rows indicate the source model used to optimize the SPOOK image; columns indicate the evaluation model.
All results use ImageNet targets, white initialization, and $1000$-target retrieval.
}
\label{tab:transfer}
\setlength{\tabcolsep}{5pt}
\renewcommand{\arraystretch}{1.12}
\begin{tabular}{llccc}
\toprule
Source model
& Evaluation model
& ASR@1 $\uparrow$
& ASR@5 $\uparrow$
& Median rank $\downarrow$ \\
\midrule
CLIP ViT-B/16
& CLIP ViT-B/16
& 93.9\%
& 98.4\%
& 1.0 \\
CLIP ViT-B/16
& DINOv3 ViT-B/16
& 0.6\%
& 2.1\%
& 386.0 \\
DINOv3 ViT-B/16
& DINOv3 ViT-B/16
& 85.5\%
& 94.5\%
& 1.0 \\
DINOv3 ViT-B/16
& CLIP ViT-B/16
& 0.2\%
& 0.8\%
& 449.0 \\
\midrule
CLIP ViT-B/16
& CLIP ViT-B/32
& 1.1\%
& 3.9\%
& 176.0 \\
CLIP ViT-B/32
& CLIP ViT-B/16
& 0.4\%
& 2.0\%
& 321.0 \\
\bottomrule
\end{tabular}

\vspace{1mm}
\begin{flushleft}
\footnotesize
Transfer is measured by optimizing SPOOK images on the source encoder and evaluating retrieval against the target bank of the evaluation encoder.
Median ranks are computed among $1000$ targets.
\end{flushleft}
\end{table*}

\begin{table*}[t]
\centering
\caption{
Robustness smoke test for main-table SPOOK images under simple image transformations.
Scores are target cosine similarities after transformation. ASR and rank are computed against the same 1000-target bank as the corresponding main-table setting. Each row uses a fixed random subset of 64 targets.
}
\label{tab:spook_robustness}
\small
\resizebox{\textwidth}{!}{
\begin{tabular}{llrrrrrrr}
\toprule
Condition & Transform & Score & $\Delta$ Score & ASR@1 & $\Delta$ ASR@1 & ASR@5 & $\Delta$ ASR@5 & Med. Rank \\
\midrule
CLIP / ImageNet & None & 0.3355 & 0.0000 & 93.8 & 0.0 & 98.4 & 0.0 & 1 \\
CLIP / ImageNet & JPEG quality 75 & 0.2296 & 0.1059 & 14.1 & 79.7 & 31.2 & 67.2 & 23 \\
CLIP / ImageNet & Center crop 90\% & 0.2024 & 0.1331 & 4.7 & 89.1 & 14.1 & 84.3 & 119 \\
CLIP / ImageNet & Center crop 80\% & 0.1922 & 0.1433 & 4.7 & 89.1 & 12.5 & 85.9 & 162 \\
CLIP / ImageNet & Gaussian blur $r=1$ & 0.2189 & 0.1166 & 18.8 & 75.0 & 34.4 & 64.0 & 21 \\
CLIP / ImageNet & Resize 112 then restore & 0.2149 & 0.1206 & 14.1 & 79.7 & 23.4 & 75.0 & 21 \\
\midrule
CLIP / Flickr30k & None & 0.3518 & 0.0000 & 82.8 & 0.0 & 95.3 & 0.0 & 1 \\
CLIP / Flickr30k & JPEG quality 75 & 0.2303 & 0.1215 & 3.1 & 79.7 & 15.6 & 79.7 & 121 \\
CLIP / Flickr30k & Center crop 90\% & 0.1952 & 0.1566 & 0.0 & 82.8 & 7.8 & 87.5 & 129 \\
CLIP / Flickr30k & Center crop 80\% & 0.1897 & 0.1621 & 0.0 & 82.8 & 4.7 & 90.6 & 225 \\
CLIP / Flickr30k & Gaussian blur $r=1$ & 0.2234 & 0.1284 & 3.1 & 79.7 & 6.2 & 89.1 & 92 \\
CLIP / Flickr30k & Resize 112 then restore & 0.2157 & 0.1360 & 1.6 & 81.2 & 7.8 & 87.5 & 92 \\
\midrule
DINOv3 / ImageNet & None & 0.5266 & 0.0000 & 82.8 & 0.0 & 93.8 & 0.0 & 1 \\
DINOv3 / ImageNet & JPEG quality 75 & -0.0319 & 0.5586 & 0.0 & 82.8 & 1.6 & 92.2 & 238 \\
DINOv3 / ImageNet & Center crop 90\% & 0.0141 & 0.5126 & 4.7 & 78.1 & 7.8 & 86.0 & 159 \\
DINOv3 / ImageNet & Center crop 80\% & 0.0026 & 0.5240 & 1.6 & 81.2 & 3.1 & 90.7 & 216 \\
DINOv3 / ImageNet & Gaussian blur $r=1$ & 0.0708 & 0.4558 & 15.6 & 67.2 & 29.7 & 64.1 & 67 \\
DINOv3 / ImageNet & Resize 112 then restore & 0.0531 & 0.4735 & 9.4 & 73.4 & 26.6 & 67.2 & 79 \\
\bottomrule
\end{tabular}
}
\end{table*}

\clearpage
\clearpage 
\bibliographystyle{IEEEtran}
    \bibliography{main}
\end{document}